\documentclass[sigconf,nonacm]{acmart}

\usepackage{booktabs}
\usepackage{multirow}
\usepackage{amsmath}
\usepackage{algorithm}
\usepackage{algorithmic}

\graphicspath{{figures/}}

\providecommand{\checkmark}{\ding{51}}
\newcommand{\best}[1]{\begingroup\setlength{\fboxsep}{0.45pt}\colorbox{black!15}{\strut\textbf{#1}}\endgroup}
\newcommand{\na}{\textemdash{}}

\title{GeoPAR: Large-Scale Multi-Agent Combinatorial Optimization with Geometry-Guided Parallel Autoregressive Learning}

\author{Wenjian Wu}
\email{wjwuwwj@stu.suda.edu.cn}
\affiliation{%
  \department{School of Future Science and Engineering,}
  \institution{Soochow University}
  \city{Suzhou}
  \country{China}
}

\author{Zesheng Jia}
\email{zsjia@stu.suda.edu.cn}
\affiliation{%
  \department{School of Future Science and Engineering,}
  \institution{Soochow University}
  \city{Suzhou}
  \country{China}
}

\author{Jiaying Tang}
\email{jytang0922@stu.suda.edu.cn}
\affiliation{%
  \department{School of Future Science and Engineering,}
  \institution{Soochow University}
  \city{Suzhou}
  \country{China}
}

\author{Benyuan Yang}
\authornote{Corresponding Author.}
\email{byyang@suda.edu.cn}
\affiliation{%
  \department{School of Future Science and Engineering,}
  \institution{Soochow University}
  \city{Suzhou}
  \country{China}
}

\author{Jin Wang}
\authornotemark[1]
\email{wjin1985@suda.edu.cn}
\affiliation{%
  \department{School of Future Science and Engineering,}
  \institution{Soochow University}
  \city{Suzhou}
  \country{China}
}

\begin{document}

\begin{abstract}
Multi-agent combinatorial optimization problems are notoriously challenging due to their NP-hard nature. Recent parallel autoregressive neural solvers improve inference efficiency by allowing agents to make decisions simultaneously, but their performance often degrades on large-scale instances. This is largely attributable to weak modeling of local geometric structures and the fact that conflicting task selections are handled only after action generation. To address these limitations, we propose GeoPAR, a geometry-guided parallel autoregressive reinforcement learning framework for scalable multi-agent combinatorial optimization. GeoPAR integrates three key components: (1) a projection-window sparse geometry mechanism that builds lightweight local candidate neighborhoods through multi-directional projections, (2) sparse edge-biased attention that injects these geometric relations into node representations, and (3) cache-guided conflict-aware assignment that reuses the geometric cache during decoding to suppress duplicate selections of exclusive tasks. Experiments on heterogeneous vehicle routing and open multi-depot pickup-and-delivery problems show that GeoPAR improves large-scale zero-shot generalization while substantially reducing rollout steps and maintaining efficient inference.
\end{abstract}



\maketitle

\section{Introduction}
\begin{figure}[!t]
	\centering
	\includegraphics[width=\columnwidth, trim=0.7cm 0.7cm 0.7cm 0.7cm, clip]{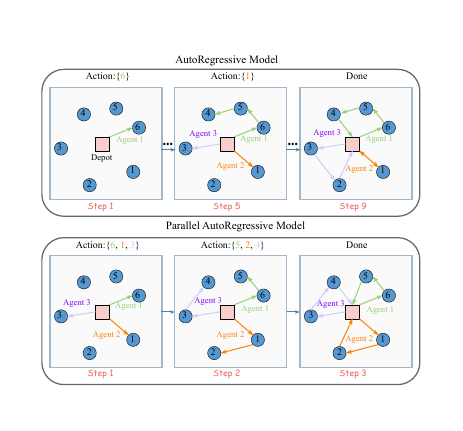}
    \caption{From sequential to parallel autoregressive models.}
    \label{MDP}
\end{figure}

Multi-agent combinatorial optimization problems are commonly formulated as optimization problems over discrete spaces, with the goal of generating high-quality action and route sequences for a group of agents~\cite{liao2025bopo}. These problems arise widely in logistics and supply-chain scheduling~\cite{Weinand2022ResearchTrends}, multi-robot task allocation~\cite{Chen2025MultiUAV}, vehicle routing~\cite{friggstad2026breaching}, and dynamic service-system optimization~\cite{Qian2024MGNet}. In many scenarios, feasible solutions must be constructed by coordinating multiple heterogeneous agents with distinct attributes. Therefore, efficient coordination of heterogeneous agents under complex constraints remains a central challenge in multi-agent combinatorial optimization~\cite{ka2024systematic}.

Many representative multi-agent combinatorial optimization problems are NP-hard~\cite{ausiello2012complexity, Peres2021Metaheuristics, wu2025efficient, yang2026systematic}. Consequently, finding globally optimal solutions within practical time limits is often difficult. Although a range of exact and heuristic methods have been developed to address these problems~\cite{Babincsak2023AntColonyMRTA, Shan2024DistributedMRTA, lu2024digital}, conventional solvers typically incur substantial computational costs and depend on hand-crafted heuristic rules that are closely tied to specific problem settings.

In recent years, neural combinatorial optimization (NCO) has offered a promising alternative by learning neural construction policies that produce near-optimal assignment decisions in a short time~\cite{Li2025CaDA,Luo2023LEHD,Zong2022MAPDP}. In particular, reinforcement learning-based NCO methods optimize policies through interaction with the environment without relying on manually labeled optimal solutions, and the learned policies can match or even outperform conventional methods~\cite{Yi2026RADAR,Liu2026ScaleNet}. Within the NCO framework, autoregressive (AR) models are widely adopted because they construct feasible solutions step by step~\cite{Dai2025HetMRTA,Li2025AGVWarehouse}. However, as the task scale increases, serial decoding incurs substantial time costs and limits the efficiency of parallel decision-making among multiple agents. To address this limitation, recent studies have proposed parallel autoregressive combinatorial optimization (PA-CO) frameworks~\citep{Berto2025PARCO}. As illustrated in Figure~\ref{MDP}, these frameworks allow multiple agents to generate actions simultaneously at the same construction step.

Although PA-CO shows promising efficiency and solution quality, zero-shot generalization to large-scale instances still faces two key challenges. First, when the number of tasks increases substantially, the policy must handle not only a larger action space but also shifted local geometric structures~\cite{Liu2026ScaleNet,Chen2025TTPL,Luo2023LEHD}. Consequently, a policy network trained on small-scale graphs may fail to preserve effective local successor structures on large-scale graphs. The model can still perform parallel decoding in form, but the quality of the generated decisions may become unstable under scale shifts. Second, PA-CO methods typically allow multiple agents to generate action proposals independently and then repair conflicting selections through a conflict-handling mechanism. This design may repeatedly assign mutually exclusive tasks to different agents, while conflicts are resolved only after action generation~\cite{Berto2025PARCO,Zang2025Plasticity}. As a result, the training signal provides limited direct guidance for learning conflict-aware decision making during action generation.

\begin{figure}[!h]
\centering
\includegraphics[width=\columnwidth]{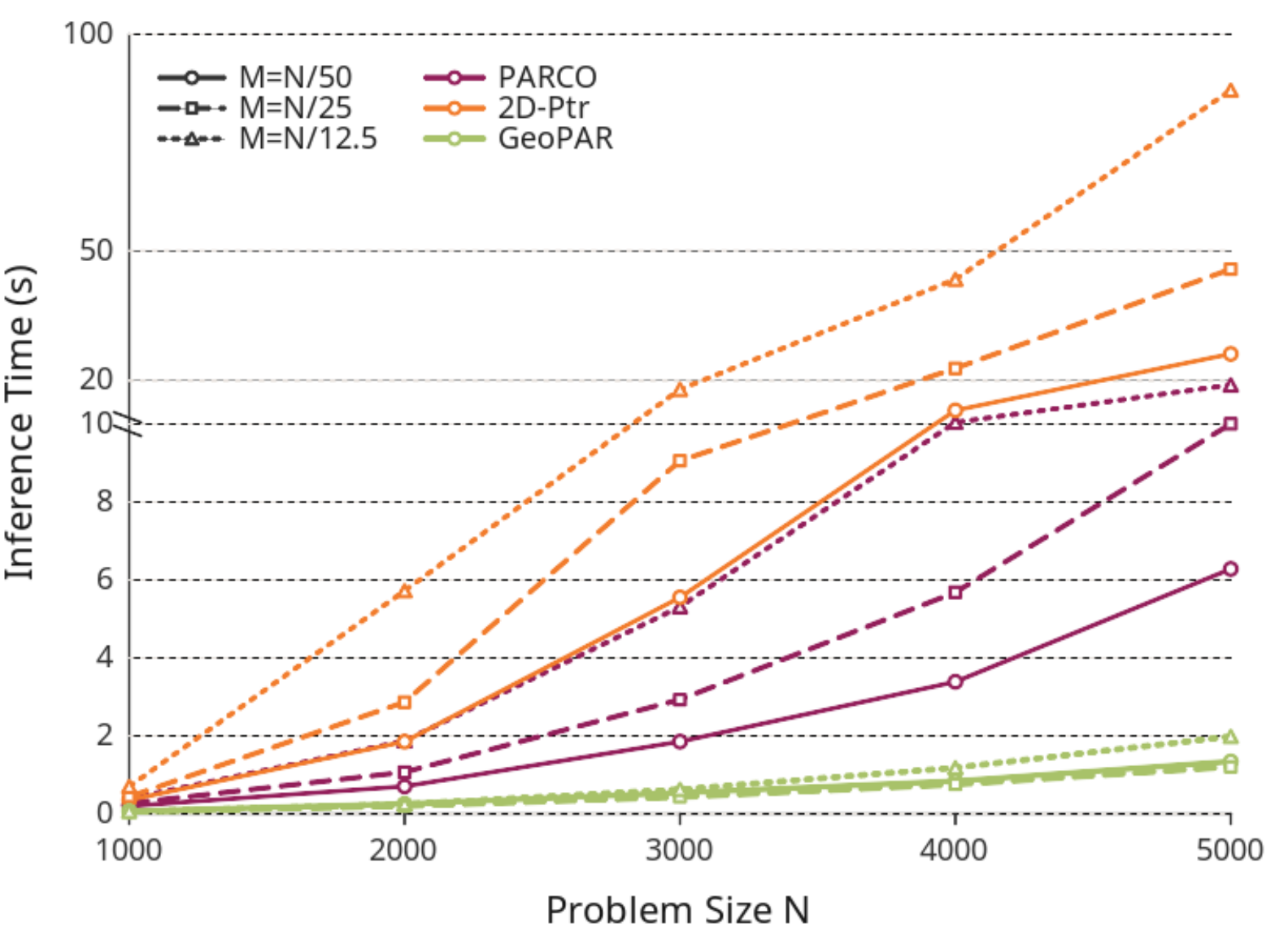}
\caption{Inference-time speedups of GeoPAR versus PARCO and 2D-Ptr on large-scale instances. Compared with the PARCO and 2D-Ptr baselines, GeoPAR achieves average inference-time speedups of $7.5\times$ and $33.8\times$. In our experiments, we set the number of agents $M$ as $N/k$ to maintain a constant agent density of one agent per $k$ tasks, where $N$ is the number of tasks.}
\label{fig:hcvrp_inference_scaling}
\end{figure}

To address these challenges, we propose GeoPAR, a geometry-guided parallel autoregressive policy. GeoPAR constructs sparse geometric candidate neighborhoods through a projection-window (PWin) mechanism and captures local transition structures during encoding using sparse edge-biased attention. During decoding, the model reuses this geometric cache and integrates neural pointer scores, geometric candidates, and feasibility signals to construct compact candidate pools. It then reduces duplicate selections among simultaneous actions through conflict-aware (CA) parallel assignment. By integrating local geometric modeling with CA assignment, GeoPAR maintains efficient parallel inference while improving zero-shot generalization from small training instances to larger ones. This claim is supported by the empirical results in Figure~\ref{fig:hcvrp_inference_scaling}. Our contributions are summarized as follows:
\begin{itemize}
    \item We design a projection-window sparse geometry mechanism that constructs lightweight local candidate neighborhoods through multi-directional projections, enabling efficient modeling of local transition structure among tasks.
    \item We propose cache-guided conflict-aware parallel assignment, which reuses the geometric cache during decoding and suppresses duplicate task selections in simultaneous agent actions.
    \item We evaluate GeoPAR on multiple multi-agent combinatorial optimization tasks and show that it provides favorable large-scale generalization in solution quality, rollout steps, and inference time.
\end{itemize}

%================================================================
\section{Related Work}
\paragraph{Multi-agent AR methods for NCO}
NCO typically formulates task allocation as a sequential construction process, in which an AR decoder selects nodes, tasks, or actions step by step~\citep{Kwon2020POMO,Kwon2021MatNet,Zheng2024DPN}. Because AR models naturally incorporate dynamic state updates and feasibility masks, they have become a dominant framework for neural routing solvers~\citep{Liu20242DPtr,smit2026neural,strang2026planning}. Although these models are more effective than purely decentralized policies for multi-agent routing variants~\citep{Park2023MinMaxMTSP,wang2024dyps}, their sequential structure still leads to high generation latency. Recent parallel AR methods introduce multiple pointers, agent communication, and conflict-handling mechanisms, allowing actions to be executed in parallel within the same construction step~\citep{Berto2025PARCO}. However, existing methods mainly correct duplicate or incompatible selections after candidate generation, while local structures and exclusivity constraints are not fully considered when candidate actions are constructed. In contrast, GeoPAR incorporates projection-window-based geometric candidates and cache-guided conflict-aware assignment into parallel AR decoding, enabling local structure and conflict awareness to jointly guide synchronous multi-agent decisions.

\paragraph{Large-scale NCO methods}
Generalization from small training instances to large test instances remains a key challenge in NCO~\citep{Liu2026ScaleNet}. Existing methods for large-scale generalization can be broadly grouped into three categories. Adaptation-based methods mitigate scale gaps through test-stage projection, increased decoder computation, or instance-specific adjustment, but these designs introduce additional inference costs~\citep{Chen2025TTPL,Luo2023LEHD}. Divide-and-conquer methods decompose large instances into subproblems and merge the resulting solutions, but their performance largely depends on the partitioning and merging strategies~\citep{Zheng2024UDC,Pan2025HLGP}. Local search and improvement methods use neural models to guide neighborhood refinement or large neighborhood search, but they usually require multiple rounds of iterative refinement and may rely on problem-specific designs~\citep{Hottung2022EAS,Huang2023CLLNS,Li2025DRHG}. In contrast to these approaches, GeoPAR focuses on single-pass parallel construction. It constructs a reusable sparse geometric cache before decoding and reuses this cache during conflict-aware assignment, thereby improving the effectiveness of each parallel action without relying on test-time fine-tuning or repeated local search.

%================================================================
\section{Preliminaries}

\subsection{Collaborative Multi-Agent MDP}
We formulate the problem as a collaborative multi-agent MDP
\(\langle \mathcal{S}, \lbrace\mathcal{A}_m\rbrace_{m=1}^{M}, P, R, \kappa\rangle\),
where \(\mathcal{S}\) is the state space, \(M\) is the number of agents, and \(m\in\{1,\ldots,M\}\) indexes an agent. The feasible action set \(\mathcal{A}_m(s_t)\) of agent \(m\) includes task selections, waiting, and returning to a depot; \(P\) is the transition kernel, \(R\) is the reward function, and \(\kappa\) is a conflict-handling function. At decision step $t$, the state of the environment is denoted by $s_t$. Each agent selects one action, forming the joint action
$\mathbf{a}_t = (a_t^m)_{m=1}^{M}$, where $a_t^m\in\mathcal{A}_m(s_t)$.

The policy maps the current state to a joint action distribution:
$\pi_{\theta}(\mathbf{a}_t\mid s_t)$, where $\mathbf{a}_t\in
\prod_{m=1}^{M}\mathcal{A}_m(s_t)$. Since multiple agents may select mutually conflicting actions, we apply the conflict handling function
\begin{equation}
\kappa:
\left(\prod_{m=1}^{M}\mathcal{A}_m\right)\times\mathcal{S}
\rightarrow
\prod_{m=1}^{M}\mathcal{A}_m .
\end{equation}

The executable joint action is therefore $\tilde{\mathbf{a}}_t=\kappa(\mathbf{a}_t,s_t)$, where duplicated assignments are removed or replaced by waiting. The environment then evolves according to $s_{t+1} =P(s_t \mid \tilde{\mathbf{a}}_t)$,
and the policy receives reward $r_t =R(s_t\mid \tilde{\mathbf{a}}_t)$.
The final solution is represented by the conflict free joint action sequence
$\tilde{\mathbf{a}}_{1:T} = (\tilde{\mathbf{a}}_1,\ldots,\tilde{\mathbf{a}}_T)$.

\subsection{Parallel Autoregressive Models}

Different from AR models that decode actions one by one, a PA model decodes the actions of multiple agents in parallel within each step. Let \(h=f_{\theta}(s_0)\) be the encoded representation of the initial problem state. At each step \(t\), the joint action proposal is generated in parallel over available agents. Let \(\mathcal{M}_t\) denote the set of available agents at state \(s_t\). The joint proposal is factorized as
\begin{equation}
\pi_{\theta}(\mathbf{a}_t \mid \tilde{\mathbf{a}}_{<t}, s_t, h)
=
\prod_{m\in \mathcal{M}_t}
\pi_{\theta}(a_t^m \mid \tilde{\mathbf{a}}_{<t}, s_t, h, m).
\end{equation}
%where \(a_t^m\in\mathcal{A}_m(s_t)\) is the action proposed for agent \(m\).
Because these actions are decoded simultaneously, the raw proposal \(\mathbf{a}_t\) may still contain conflicts. Therefore, we use the conflict-handling function $\tilde{\mathbf{a}}_t=\kappa(\mathbf{a}_t,s_t)$. Thus, the model remains autoregressive across decision steps without imposing an arbitrary sequential ordering on agents within each step.

%=======================================================================
\section{Method: GeoPAR}\label{sec:method_geopar}

\begin{figure*}[t]
	\centering
	\includegraphics[width=\textwidth, trim=0.7cm 0.5cm 0.7cm 0.45cm, clip]{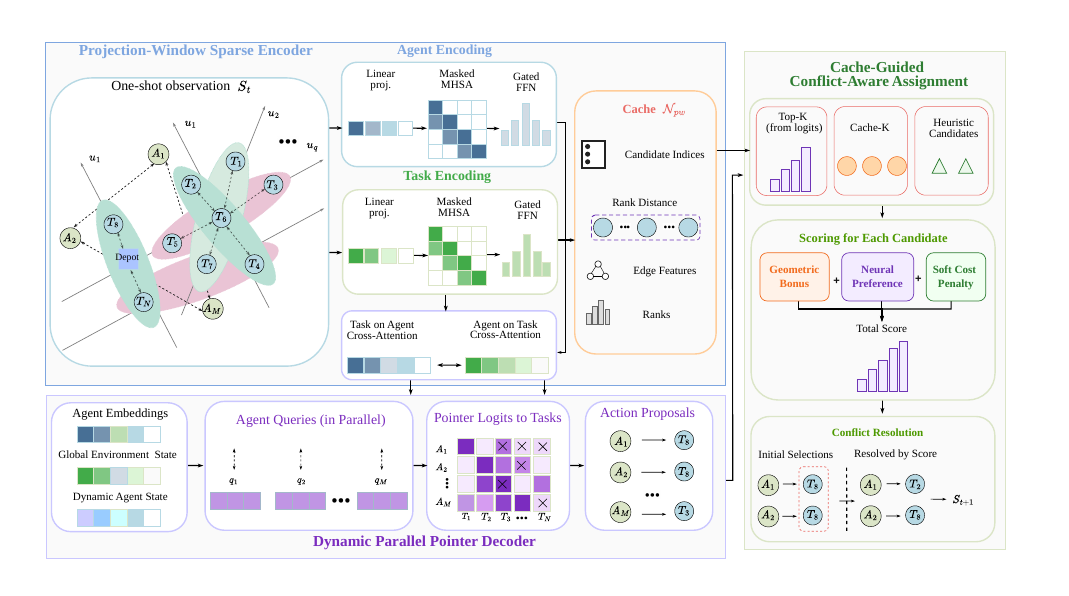}  %  [width=\textwidth]
	%	\hfillImpro
    \caption{Overview of GeoPAR. The projection-window sparse encoder maps heterogeneous agents and task nodes into a shared representation space. Task nodes are sorted along multiple projection directions to construct a reusable geometric cache. The encoder refines node and agent embeddings through sparse edge-biased attention and agent-task interaction. At each decision step, the dynamic parallel pointer decoder forms agent queries from encoded embeddings, dynamic agent states, and the global environment state, producing simultaneous action logits. Finally, the cache-guided conflict-aware assignment module outputs a feasible joint action.}\label{model}
\end{figure*}

We propose GeoPAR, a geometry-aware parallel autoregressive policy for scalable multi-agent combinatorial optimization, as illustrated in Figure~\ref{model}. At each decoding step, multiple heterogeneous agents generate actions simultaneously, while a parallel assignment operator enforces feasibility. GeoPAR projects agents and tasks into a shared embedding space and constructs a lightweight geometric cache to represent local transition structures. The decoder combines these representations with dynamic agent queries to score actions in parallel, and the assignment module converts the result into feasible actions.

\subsection{Projection-Window Sparse Encoder}\label{subsec}

\paragraph{Heterogeneous input embedding.}
Given an instance $x$ with $M$ agents and $N$ task nodes, the encoder first maps their raw features into a shared $d$-dimensional embedding space. Let $X_A \in \mathbb{R}^{M \times k_A}, X_V \in \mathbb{R}^{N \times k_V}$
denote the agent feature matrix and the node feature matrix, where $k_A$ and $k_V$ are the corresponding feature dimensions. Since agents and task nodes have different semantics and feature spaces, we use two separate linear projections:
$W_A \in \mathbb{R}^{k_A \times d}, W_V \in \mathbb{R}^{k_V \times d}$.
 The initial agent and node embeddings are then defined as
$H_A^{(0)} = X_A W_A, H_V^{(0)} = X_V W_V$,
where $H_A^{(0)} \in \mathbb{R}^{M \times d}$ and $H_V^{(0)} \in \mathbb{R}^{N \times d}$. We concatenate the two sets of embeddings as
$H^{(0)} = [H_A^{(0)}; H_V^{(0)}]$,
which is used as the input to the subsequent geometry-aware encoder layers. This separate embedding design preserves the heterogeneity between agents and task nodes while projecting them into a unified representation space for later interaction modeling.

\paragraph{Projection-window cache.}
To avoid dense node-node relations, we construct a sparse geometric cache by sorting task nodes under multiple one-dimensional projection orderings. Let $z_i\in\mathbb{R}^{p}$ denote the projection descriptor of task node $i$, which can be instantiated by its normalized coordinate or other lightweight geometric features. Let $\lbrace u_\ell\rbrace_{\ell=1}^{q}$ be $q$ unit projection directions with $u_\ell\in\mathbb{R}^{p}$. Each direction defines an ordering key
$\eta_\ell(i)=z_i^\top u_\ell$.

We denote the set of ordering functions by
$
\mathcal{R}=\lbrace\eta_\ell\rbrace_{\ell=1}^{q}\cup\mathcal{R}_{\mathrm{aux}}$,
where $\mathcal{R}_{\mathrm{aux}}$ contains optional task-native orderings, such as depot-distance ordering for routing or pair-aware ordering for pickup-delivery problems.

For each $\eta\in\mathcal{R}$, let $r_{\eta}(i)$ be the position of node $i$ after sorting all task nodes by $\eta(i)$. Given a window radius $w$, the projection window for node $i$ is defined as
\begin{equation}
\begin{aligned}
\widehat{\mathcal{N}}_{\mathrm{pw}}(i)
&=
\{i\}\cup
\left\{
j:\exists\,\eta\in\mathcal{R}
\ \text{with}\
|r_{\eta}(j)-r_{\eta}(i)|\le w
\right\},
\\
\mathcal{N}_{\mathrm{pw}}(i)
&=
\operatorname{TopK}
\left(
\widehat{\mathcal{N}}_{\mathrm{pw}}(i), K
\right).
\end{aligned}
\end{equation}

Here $\widehat{\mathcal{N}}_{\mathrm{pw}}(i)$ is the merged candidate set before truncation. $\operatorname{TopK}(\cdot,K)$ keeps at most $K$ candidates with the smallest rank distance to node $i$. Before truncation, each node has at most $1+2w|\mathcal{R}|$ candidates.

\paragraph{Sparse edge-biased enhancement.}
The collection of projection-window neighborhoods
$\mathcal{N}_{\mathrm{pw}}=\lbrace\mathcal{N}_{\mathrm{pw}}(i)\rbrace_{i=1}^{N}$
serves as the sparse geometric cache used by the encoder. Given this cache, the encoder refines node embeddings by restricting node-node attention to cached neighborhoods. For node $i$, let $\mathcal{N}_i=\mathcal{N}_{\mathrm{pw}}(i)$ denote its cached candidate set. For each cached pair $(i,j)$ with $j\in\mathcal{N}_i$, we compute an edge descriptor $\psi_{ij}$, a candidate-source type $c_{ij}$, and a normalized rank distance $\rho_{ij}$ from the projection-window construction. For each attention head $r$, the score from node $i$ to candidate node $j$ is computed as
\begin{equation}
\begin{aligned}
e_{ij}^{(r)}
&=
a_{ij}^{(r)}
+
b_{\psi}^{(r)}(\psi_{ij})
+
b_c^{(r)}(c_{ij})
+
b_{\rho}^{(r)}(\rho_{ij}),
\end{aligned}
\end{equation}
where $a_{ij}^{(r)}$ is the standard scaled dot-product attention score computed from the normalized node embeddings. The bias terms $b_{\psi}^{(r)}$, $b_c^{(r)}$, and $b_{\rho}^{(r)}$ inject edge geometry, candidate-source information, and ranking proximity into the sparse attention score.

The scores are normalized over $j\in\mathcal{N}_i$ and used to aggregate local candidate features, producing $g_i^{\mathrm{loc}}$. In parallel, node $i$ attends to the agent embeddings $H_A$ to obtain an agent-conditioned update $g_i^{\mathrm{agt}}$. The final node representation is updated by
\begin{equation}
\tilde h_i^V
=
h_i^V
+
\gamma\odot
\mathrm{LN}
\left(
W_F[\,g_i^{\mathrm{loc}};g_i^{\mathrm{agt}}\,]
\right),
\end{equation}
where $W_F$ is a fusion projection, $\gamma$ is a learnable residual gate, $\odot$ denotes element-wise multiplication, and $[\cdot;\cdot]$ denotes concatenation. The enhanced node embeddings are then passed to subsequent encoder layers, while $\mathcal{N}_{\mathrm{pw}}$ is retained for decoder-side assignment.

\subsection{Dynamic Parallel Pointer Decoder}\label{subsec:dynamic_parallel_decoder}

Given the encoder output $H=[H_A;H_V]$, GeoPAR constructs solutions in a parallel autoregressive manner. At each step $t$, the decoder builds one dynamic query for each active agent by combining its static embedding with the current agent and environment states:
\begin{equation}
q_t^m
=
W_q
\left[
h_m^A;\,
\phi_A(\delta_t^m);\,
\phi_E(s_t)
\right],
\qquad m\in\mathcal{M}_t,
\end{equation}
where $\mathcal{M}_t$ is the set of active agents, $h_m^A$ is the encoded embedding of agent $m$, $\delta_t^m$ denotes its dynamic state, and $\phi_E(s_t)$ summarizes the global environment state. The agent queries are then passed through a lightweight communication block,
\begin{equation}
\bar Q_t=\mathrm{Comm}_{\theta}(Q_t),
\qquad
Q_t={\left[q_t^m\right]}_{m\in\mathcal{M}_t},
\end{equation}
so that active agents can exchange information before action scoring. The communicated queries attend to the encoded action embeddings under the feasibility mask $M_t$. Let $\mathcal{A}_t$ denote the current action set, including unserved task actions and task-native safe actions such as waiting or returning to a depot. For each action $j\in\mathcal{A}_t$, we define its dynamic action embedding as
\[
\bar h_t^j=h^j+W_{\xi}\xi_t^j,
\]
where $h^j$ is the static embedding of action $j$ and $\xi_t^j$ is its dynamic feature at step $t$. The pointer logit of agent $m$ choosing action $j$ is computed as
\begin{equation}
L_t^{m,j}
=
\beta\cdot
\tanh
\left(
\frac{
\langle
W_Q\bar q_t^m,
W_K\bar h_t^j
\rangle
}{\sqrt{d}}
\right).
\end{equation}
Hard-infeasible actions are masked by setting $L_t^{m,j}=-\infty$ when $M_t^{m,j}=0$. This yields the per-agent neural action distribution
\begin{equation}
\pi_{\theta}(a_t^m=j\mid s_t,H)
=
{\mathrm{Softmax}\left(L_t^{m,\cdot}\right)}_j .
\end{equation}

The decoder therefore produces parallel per-agent action scores from the same state $s_t$. Since independently selected high-scoring actions may still conflict over exclusive tasks, GeoPAR passes $L_t$, $M_t$, $s_t$, and the projection-window cache $\mathcal{N}_{\mathrm{pw}}$ to the cache-guided assignment module, which returns the executable joint action $\tilde{\mathbf{a}}_t$ used for the environment transition.

\subsection{Cache-Guided Conflict-Aware Assignment}\label{subsec:cache_guided_assignment}

For each agent $m$, GeoPAR constructs a compact candidate pool:
$\mathcal{B}_t^m=
\mathrm{TopK}_{\mathrm{logit}}(m)
\cup
\mathrm{CacheK}(m,\mathcal{N}_{\mathrm{pw}},s_t)
\cup
\mathrm{HeurK}(m,s_t)$,
where the three terms correspond to high-probability decoder actions, cache-based geometric successors, and lightweight task-native candidates, respectively. During training, a few random feasible actions are also added to encourage exploration.

Each candidate $j\in\mathcal{B}_t^m$ is then scored by combining neural preference, geometric guidance, and immediate soft costs:
\begin{equation}
\label{eq:assignment_score}
S_t^{m,j}
=
\ell_{\theta,t}^{m,j}
+
\lambda_{\mathrm{geo}} B_{\mathrm{geo}}^{m,j}
-
\Omega_t^{m,j},
\end{equation}
where $\ell_{\theta,t}^{m,j}=\log \pi_{\theta}(j\mid s_t,H,m)$ denotes the decoder score, $B_{\mathrm{geo}}^{m,j}$ is the cache-derived geometric bonus, and $\Omega_t^{m,j}$ penalizes undesirable immediate effects such as travel cost, estimated makespan increase, or load imbalance. Hard infeasibility, such as capacity violation, visited-task reuse, or precedence violation, is handled by the feasibility mask $M_t$.

Based on these scores, the assignment module converts parallel per-agent scores into a feasible action. The key constraint is that exclusive actions cannot be assigned to multiple agents in the same parallel step. To enforce this, conflict-aware sampling or resolution maintains an internal consumed set $\mathcal{D}_t$ of accepted exclusive actions. During training, agents sample from feasible candidates while excluding consumed exclusive actions. During inference, agents submit ranked proposals, and each conflicted exclusive action is assigned only to the agent with the highest score.

Algorithm~\ref{alg:cache_guided_assignment} summarizes the above process using five operators: $\operatorname{Pool}$ builds $\mathcal{B}_t^m$, $\operatorname{Score}$ computes Eq.~(\ref{eq:assignment_score}), $\operatorname{TopR}$ keeps ranked proposals, $\operatorname{Sample}_{\mathcal{X}}$ performs training-time conflict-aware sampling, and $\operatorname{Resolve}_{\mathcal{X}}$ performs inference-time conflict resolution over exclusive actions.

\begin{algorithm}[t]
\caption{Cache-Guided Parallel Assignment}\label{alg:cache_guided_assignment}
\begin{algorithmic}[1]
\REQUIRE{} Logits $L_t$, mask $M_t$, state $s_t$, cache $\mathcal{N}_{\mathrm{pw}}$, exclusive set $\mathcal{X}$, mode $o$
\ENSURE{} Feasible joint action $\tilde{\mathbf{a}}_t$
\STATE{} $\mathcal{B}_t^m \leftarrow \operatorname{Pool}(m,L_t,s_t,\mathcal{N}_{\mathrm{pw}}),\quad \forall m$
\STATE{} $S_t^{m,j} \leftarrow \operatorname{Score}(m,j,L_t,s_t,\mathcal{N}_{\mathrm{pw}}),\quad \forall m,\ j\in\mathcal{B}_t^m$
\STATE{} $\mathcal{P}_t^m \leftarrow \operatorname{TopR}(\mathcal{B}_t^m,S_t^m),\quad \forall m$
\IF{$o=\textsc{train}$}
    \STATE{} $\tilde{\mathbf{a}}_t \leftarrow \operatorname{Sample}_{\mathcal{X}}(\lbrace\mathcal{P}_t^m\rbrace_{m=1}^{M}, M_t)$
\ELSE{}
    \STATE{} $\tilde{\mathbf{a}}_t \leftarrow \operatorname{Resolve}_{\mathcal{X}}(\lbrace\mathcal{P}_t^m\rbrace_{m=1}^{M}, S_t, M_t)$
\ENDIF{}
\STATE{} $\tilde{\mathbf{a}}_t \leftarrow \operatorname{Fallback}(\tilde{\mathbf{a}}_t,M_t)$
\RETURN{} $\tilde{\mathbf{a}}_t$
\end{algorithmic}
\end{algorithm}

Since $\operatorname{Sample}_{\mathcal{X}}$ and $\operatorname{Resolve}_{\mathcal{X}}$ both reject already consumed exclusive actions, the resulting joint action contains no duplicate exclusive action within each parallel step. Under the feasibility mask and the safe-fallback assumption, the returned action is feasible for environment execution.

\subsection{Training Objective}

GeoPAR is trained with reinforcement learning. Let $R(\tilde{\mathbf{a}}_{1:T})$ be the terminal reward, typically the negative task cost. The objective is
\begin{equation}
    J(\theta)
    =
    \mathbb{E}_{\tilde{\mathbf{a}}_{1:T}\sim\pi_{\theta}^{\mathrm{train}}}
    \left[
    R(\tilde{\mathbf{a}}_{1:T})
    \right],
\end{equation}
where $\pi_{\theta}^{\mathrm{train}}$ denotes the rollout distribution induced by the decoder scores and the training-time conflict-aware sampling operator $\operatorname{Sample}_{\mathcal{X}}$. At step $t$, the log-probability used for policy-gradient training is the probability of the executable joint action sampled by $\operatorname{Sample}_{\mathcal{X}}$. Let $\mathcal{E}_t^m$ denote the feasible candidate set of agent $m$ after applying the feasibility mask and removing already consumed exclusive actions. Then
\begin{equation}
\log \pi_{\theta}^{\mathrm{train}}(\tilde{\mathbf{a}}_t\mid s_t)
=
\sum_{m\in\mathcal{M}_t}
\log
\frac{
\exp(S_t^{m,\tilde a_t^m}/\tau)
}{
\sum_{j\in\mathcal{E}_t^m}
\exp(S_t^{m,j}/\tau)
},
\end{equation}
where $S_t^{m,j}$ is the assignment score in Eq.~(\ref{eq:assignment_score}) and $\tau$ is the sampling temperature. With an action-independent baseline $b$, the score-function estimator is
\begin{equation}
    \nabla_{\theta}J(\theta)
    =
    \mathbb{E}
    \left[
        (R-b)
        \sum_{t=1}^{T}
        \nabla_{\theta}
        \log \pi_{\theta}^{\mathrm{train}}(\tilde{\mathbf{a}}_t\mid s_t)
    \right].
\end{equation}
At inference time, GeoPAR uses deterministic conflict resolution with the same assignment scores, and no policy-gradient probability is required for $\operatorname{Resolve}_{\mathcal{X}}$.

%=======================================================================
\section{Experiments}
\begin{table*}[!t]
\centering
\caption{Results on HCVRP and OMDCPDP. The gap is computed relative to the best objective value within each $N/M$ setting. All results were obtained from five replicate experiments. Lower values are better.} %The best objective values, zero gaps and the smallest numbers of rollout steps are highlighted.
\label{tab:hcvrp_compact_overview}
\small
\renewcommand{\arraystretch}{0.82}
\setlength{\tabcolsep}{0pt}
\begin{tabular*}{\textwidth}{@{}c@{\extracolsep{\fill}}r@{\hspace{0.22em}}r@{\hspace{0.22em}}r@{\hspace{0.32em}}r@{\extracolsep{\fill}}r@{\hspace{0.22em}}r@{\hspace{0.22em}}r@{\hspace{0.32em}}r@{\extracolsep{\fill}}r@{\hspace{0.22em}}r@{\hspace{0.22em}}r@{\hspace{0.32em}}r@{\extracolsep{\fill}}r@{\hspace{0.22em}}r@{\hspace{0.22em}}r@{\hspace{0.32em}}r@{}}
\toprule
\multicolumn{17}{c}{\textbf{Small-scale HCVRP}} \\
\midrule
N/M & \multicolumn{4}{c}{60/3} & \multicolumn{4}{c}{60/5} & \multicolumn{4}{c}{100/5} & \multicolumn{4}{c}{100/7} \\
\cmidrule(lr){2-5}\cmidrule(lr){6-9}\cmidrule(lr){10-13}\cmidrule(lr){14-17}
Method & Obj. & Gap & Steps & Time & Obj. & Gap & Steps & Time & Obj. & Gap & Steps & Time & Obj. & Gap & Steps & Time \\
\midrule
SISRs~\cite{Christiaens2020SISR} & \best{6.57} & \best{0.00\%} & \na{} & {271s} & \best{4.00} & \best{0.00\%} & \na{} & {274s} & \best{6.17} & \best{0.00\%} & \na{} & {623s} & \best{4.45} & \best{0.00\%} & \na{} & {625s} \\
GA~\cite{Karakatic2015GA} & 9.21 & 40.18\% & \na{} & 233s & 6.89 & 72.25\% & \na{} & 320s & 10.93 & 77.15\% & \na{} & 623s & 9.10 & 104.49\% & \na{} & 772s \\
SA~\cite{Ilhan2021SA} & 7.04 & 7.15\% & \na{} & 130s & 4.39 & 9.75\% & \na{} & 289s & 6.80 & 10.21\% & \na{} & 557s & 5.01 & 12.58\% & \na{} & 678s \\
AM~\cite{Kool2019Attention} & 7.41 & 12.79\% & 73.5 & \best{0.06s} & 4.59 & 14.75\% & 72.0 & \best{0.05s} & 6.96 & 12.80\% & 122.1 & \best{0.08s} & 5.12 & 15.06\% & 118.3 & 0.08s \\
2D-Ptr~\cite{Liu20242DPtr} & 7.20 & 9.59\% & 73.5 & \best{0.06s} & 4.48 & 12.00\% & 73.3 & \best{0.05s} & 6.75 & 9.40\% & 121.3 & \best{0.08s} & 4.92 & 10.56\% & 118.6 & 0.08s \\
DPN~\cite{Zheng2024DPN} & 7.66 & 16.59\% & 86.0 & \best{0.06s} & 4.93 & 23.25\% & 106.1 & 0.06s & 7.27 & 17.83\% & 154.2 & 0.09s & 5.37 & 20.67\% & 155.8 & 0.09s \\
PARCO~\cite{Berto2025PARCO} & 7.22 & 9.89\% & 58.2 & 0.08s & 4.47 & 11.75\% & 56.4 & 0.08s & 6.70 & 8.59\% & 91.2 & 0.12s & 4.87 & 9.44\% & 89.0 & 0.12s \\
GeoPAR & 7.31 & 11.26\% & \best{39.3} & 0.07s & 4.58 & 14.50\% & \best{27.1} & 0.06s & 6.82 & 10.53\% & \best{44.2} & \best{0.08s} & 4.98 & 11.91\% & \best{34.6} & \best{0.07s} \\
\midrule
\multicolumn{17}{c}{\textbf{Large-scale zero-shot HCVRP}} \\
\midrule
N/M & \multicolumn{4}{c}{1000/20} & \multicolumn{4}{c}{1000/80} & \multicolumn{4}{c}{2000/40} & \multicolumn{4}{c}{2000/160} \\
\cmidrule(lr){2-5}\cmidrule(lr){6-9}\cmidrule(lr){10-13}\cmidrule(lr){14-17}
Method & Obj. & Gap & Steps & Time & Obj. & Gap & Steps & Time & Obj. & Gap & Steps & Time & Obj. & Gap & Steps & Time \\
\midrule
AM~\cite{Kool2019Attention} & 16.03 & 16.24\% & 1187.0 & 0.35s & 4.03 & 5.77\% & 1172.0 & 0.70s & 16.17 & 14.19\% & 2361.5 & 1.84s & 4.21 & 7.67\% & 2351.0 & 5.72s \\
2D-Ptr~\cite{Liu20242DPtr} & 14.11 & 2.32\% & 1192.0 & 0.35s & 4.00 & 4.99\% & 1166.0 & 0.70s & 14.25 & 0.64\% & 2367.5 & 1.85s & 4.08 & 4.35\% & 2339.5 & 5.72s \\
DPN~\cite{Zheng2024DPN} & 29.87 & 116.61\% & 4020 & 0.60s & 4.68 & 22.83\% & 1235 & 0.20s & 26.84 & 89.55\% & 8040 & 3.14s & 4.86 & 24.30\% & 2353 & 0.99s \\
PARCO~\cite{Berto2025PARCO} & 13.97 & 1.31\% & 693.0 & 0.20s & 4.61 & 21.00\% & 562.0 & 0.39s & 15.52 & 9.60\% & 1098.0 & 0.71s & 8.35 & 113.55\% & 906.0 & 1.87s \\
GeoPAR & \best{13.79} & \best{0.00\%} & \best{215.0} & \best{0.07s} & \best{3.81} & \best{0.00\%} & \best{69.0} & \best{0.06s} & \best{14.16} & \best{0.00\%} & \best{272.0} & \best{0.26s} & \best{3.91} & \best{0.00\%} & \best{97.5} & \best{0.25s} \\
\midrule
N/M & \multicolumn{4}{c}{3000/60} & \multicolumn{4}{c}{3000/240} & \multicolumn{4}{c}{5000/200} & \multicolumn{4}{c}{5000/400} \\
\cmidrule(lr){2-5}\cmidrule(lr){6-9}\cmidrule(lr){10-13}\cmidrule(lr){14-17}
Method & Obj. & Gap & Steps & Time & Obj. & Gap & Steps & Time & Obj. & Gap & Steps & Time & Obj. & Gap & Steps & Time \\
\midrule
AM~\cite{Kool2019Attention} & 15.82 & 13.73\% & 3528.8 & 5.51s & 4.23 & 7.36\% & 3525.8 & 17.99s & 7.58 & 8.13\% & 5779.2 & 45.26s & 4.29 & 6.98\% & 5858.6 & 87.00s \\
2D-Ptr~\cite{Liu20242DPtr} & \best{13.91} & \best{0.00\%} & 3529.2 & {5.55s} & 4.03 & 2.28\% & 3510.2 & 18.01s & {7.07} & {0.85\%} & 5813.8 & {45.77s} & {4.05} & {0.99\%} & 5839.3 & {87.09s} \\
DPN~\cite{Zheng2024DPN} & 27.87 & 100.36\% & 12060 & 8.40s & 4.80 & 21.83\% & 3501 & 2.56s & 8.96 & 27.81\% & 18074 & 24.91s & 5.04 & 25.68\% & 5784 & 7.86s \\
PARCO~\cite{Berto2025PARCO} & 16.00 & 15.03\% & 1485.0 & 1.85s & 8.80 & 123.35\% & 1308.0 & 5.31s & 10.29 & 46.79\% & 1947.0 & 10.04s & 14.33 & 257.35\% & 1846.2 & 19.02s \\
GeoPAR & \best{13.91} & \best{0.00\%} & \best{291.8} & \best{0.54s} & \best{3.94} & \best{0.00\%} & \best{129.0} & \best{0.62s} & \best{7.01} & \best{0.00\%} & \best{197.6} & \best{1.21s} & \best{4.01} & \best{0.00\%} & \best{177.8} & \best{1.98s} \\
\midrule
\multicolumn{17}{c}{\textbf{Small-scale OMDCPDP}} \\
\midrule
N/M & \multicolumn{4}{c}{50/5} & \multicolumn{4}{c}{50/7} & \multicolumn{4}{c}{100/10} & \multicolumn{4}{c}{100/15} \\
\cmidrule(lr){2-5}\cmidrule(lr){6-9}\cmidrule(lr){10-13}\cmidrule(lr){14-17}
Method & Obj. & Gap & Steps & Time & Obj. & Gap & Steps & Time & Obj. & Gap & Steps & Time & Obj. & Gap & Steps & Time \\
\midrule
OR-Tools~\cite{Furnon2024ORTools} & 37.61 & 0.56\% & \na{} & 30s & \best{30.18} & \best{0.00\%} & \na{} & {30s} & 66.78 & 0.51\% & \na{} & 60s & \best{51.49} & \best{0.00\%} & \na{} & {60s} \\
HAM~\cite{Li2021HAM} & 45.16 & 20.75\% & 26.9 & 0.06s & 33.47 & 10.90\% & 18.6 & 0.03s & 78.47 & 18.11\% & 28.9 & 0.04s & 55.24 & 7.28\% & 18.5 & 0.03s \\
MAPDP~\cite{Zong2022MAPDP} & \best{37.40} & \best{0.00\%} & 16.4 & {0.05s} & 30.32 & 0.46\% & 14.2 & \best{0.02s} & 66.53 & 0.14\% & 17.6 & \best{0.02s} & 52.06 & 1.11\% & 15.4 & \best{0.02s} \\
PARCO~\cite{Berto2025PARCO} & 39.31 & 5.11\% & \best{16.3} & \best{0.02s} & 31.46 & 4.24\% & \best{14.0} & \best{0.02s} & 70.57 & 6.22\% & \best{17.0} & \best{0.02s} & 54.42 & 5.69\% & 13.3 & \best{0.02s} \\
GeoPAR & 37.93 & 1.42\% & 18.6 & 0.03s & 30.28 & 0.33\% & 14.4 & \best{0.02s} & \best{66.44} & \best{0.00\%} & 17.7 & {0.03s} & 51.57 & 0.16\% & \best{12.0} & \best{0.02s} \\
\midrule
\multicolumn{17}{c}{\textbf{Large-scale zero-shot OMDCPDP}} \\
\midrule
N/M & \multicolumn{4}{c}{1000/20} & \multicolumn{4}{c}{1000/25} & \multicolumn{4}{c}{2000/40} & \multicolumn{4}{c}{2000/50} \\
\cmidrule(lr){2-5}\cmidrule(lr){6-9}\cmidrule(lr){10-13}\cmidrule(lr){14-17}
Method & Obj. & Gap & Steps & Time & Obj. & Gap & Steps & Time & Obj. & Gap & Steps & Time & Obj. & Gap & Steps & Time \\
\midrule
OR-Tools~\cite{Furnon2024ORTools} & 2695.62 & 51.68\% & \na{} & 600s & 2232.35 & 53.19\% & \na{} & 600s & 5097.17 & 52.93\% & \na{} & 600s & 4424.12 & 61.98\% & \na{} & 600s \\
PARCO~\cite{Berto2025PARCO} & 2010.85 & 13.15\% & 91.1 & 0.11s & 1513.05 & 3.83\% & 64.9 & 0.08s & 3639.90 & 9.21\% & 92.4 & 0.12s & 2830.33 & 3.63\% & 63.3 & 0.08s \\
GeoPAR & \best{1777.15} & \best{0.00\%} & \best{54.9} & \best{0.07s} & \best{1457.22} & \best{0.00\%} & \best{44.2} & \best{0.06s} & \best{3333.04} & \best{0.00\%} & \best{55.3} & \best{0.07s} & \best{2731.29} & \best{0.00\%} & \best{45.1} & \best{0.06s} \\
\midrule
N/M & \multicolumn{4}{c}{3000/60} & \multicolumn{4}{c}{3000/75} & \multicolumn{4}{c}{4000/80} & \multicolumn{4}{c}{4000/100} \\
\cmidrule(lr){2-5}\cmidrule(lr){6-9}\cmidrule(lr){10-13}\cmidrule(lr){14-17}
Method & Obj. & Gap & Steps & Time & Obj. & Gap & Steps & Time & Obj. & Gap & Steps & Time & Obj. & Gap & Steps & Time \\
\midrule
OR-Tools~\cite{Furnon2024ORTools} & 7672.11 & 58.07\% & \na{} & 600s & 6640.27 & 66.71\% & \na{} & 600s & 10394.03 & 63.81\% & \na{} & 600s & 9107.39 & 74.94\% & \na{} & 600s \\
PARCO~\cite{Berto2025PARCO} & 5295.33 & 9.10\% & 94.3 & 0.12s & 4134.57 & 3.80\% & 61.8 & 0.08s & 6909.60 & 8.89\% & 93.6 & 0.13s & 5414.23 & 4.00\% & 61.0 & 0.09s \\
GeoPAR & \best{4853.71} & \best{0.00\%} & \best{56.0} & \best{0.07s} & \best{3983.21} & \best{0.00\%} & \best{45.8} & \best{0.06s} & \best{6345.28} & \best{0.00\%} & \best{56.4} & \best{0.08s} & \best{5205.91} & \best{0.00\%} & \best{46.2} & \best{0.06s} \\
\bottomrule
\end{tabular*}
\end{table*}

We structure the experiments to answer the following questions:
\begin{itemize}
    \item How does GeoPAR compare with conventional and neural baselines at training scales and under zero-shot scale transfer? (See Section \ref{subsec:zero_shot_scalability})
    \item Does the projection-window cache preserve task-relevant local transitions efficiently, and how sensitive is GeoPAR to the number of projection directions? (See Section \ref{subsec:projection_geometry_effect})
    \item Are projection-window geometry and conflict-aware assignment complementary? (See Section \ref{subsec:conflict_free_assignment_effect})
    \item Which components of cache-guided conflict-aware assignment drive solution quality and construction efficiency? (See Section \ref{subsec:ca_component_analysis})
    \item How does GeoPAR behave across problem families, problem scales, and agent-density regimes? (See Section \ref{subsec:robustness_density})
\end{itemize}

\subsection{Experimental Settings}
\label{subsec:experimental_settings}

\paragraph{Datasets and baselines.}
Following~\cite{Berto2025PARCO,Liu20242DPtr}, we evaluate GeoPAR on HCVRP and OMDCPDP, which model heterogeneous capacitated vehicles and open multi-depot pickup-and-delivery routing, respectively. The HCVRP baselines are SISR~\cite{Christiaens2020SISR}, GA~\cite{Karakatic2015GA}, SA~\cite{Ilhan2021SA}, AM~\cite{Kool2019Attention}, 2D-Ptr~\cite{Liu20242DPtr}, DPN~\cite{Zheng2024DPN}, and PARCO~\cite{Berto2025PARCO}; the OMDCPDP baselines are OR-Tools~\cite{Furnon2024ORTools}, HAM~\cite{Li2021HAM}, MAPDP~\cite{Zong2022MAPDP}, and PARCO.

\paragraph{Metrics.}
We report objective value, gap, rollout steps, and inference time. The HCVRP objective is the makespan, defined as the maximum speed-normalized route duration across vehicles; the OMDCPDP objective is the cumulative delivery-arrival cost. The gap is relative to the best value within each $N/M$ setting. Rollout steps measure construction length, and inference time is measured under the same GPU setting.

\paragraph{Implementation details.}
GeoPAR uses 128-dimensional policy embeddings, a projection window of size 8 with four directions, and the conflict-aware decoding configuration in Section~\ref{subsec:cache_guided_assignment}. Both models are trained for 100 epochs with a learning rate of $10^{-4}$ on small instances: 60--100 customers and 3--7 agents for HCVRP, and 50--100 tasks and 10--50 agents for OMDCPDP. Evaluation uses deterministic decoding on the instances in Table~\ref{tab:hcvrp_compact_overview}. Problem definitions and additional hyperparameters are provided in Appendices~\ref{app:problem_definitions} and~\ref{app:experimental_details}, respectively.

\subsection{Overall Performance and Zero-Shot Scalability}
\label{subsec:zero_shot_scalability}

Table~\ref{tab:hcvrp_compact_overview} presents the results for the two datasets. For HCVRP, conventional solvers and specialized neural models remain competitive on small-scale instances, where the instance size is limited and the scalability bottleneck is not yet dominant. Under large-scale zero-shot settings, AM and 2D-Ptr maintain reasonable solution quality, but they construct solutions sequentially, node by node, which causes inference time to increase sharply as $N$ grows. PARCO shortens the construction horizon through parallel decoding, but its solution quality becomes unstable. This suggests that naive parallelization may amplify coordination errors when many agents propose actions simultaneously. GeoPAR uses the projection-window cache to keep the candidate set of each agent locally meaningful as the graph size increases, while the conflict-aware assignment mechanism reduces duplicate selections in parallel decisions. Consequently, GeoPAR achieves more effective large-scale inference under zero-shot scale transfer.

For OMDCPDP, a similar pattern is observed under a different constraint structure. Although the performance differences among strong baselines are modest on small-scale instances, GeoPAR consistently achieves better performance in large-scale zero-shot settings. This indicates that the proposed mechanisms are not limited to Euclidean vehicle routing problems. The sparse geometric candidate construction provides a stable local action prior, while the assignment layer filters parallel decisions according to dynamic feasibility and task exclusivity. As the problem scale increases, OR-Tools struggles under a fixed time budget, and PARCO remains fast but exhibits unstable solution quality. In contrast, GeoPAR maintains a short construction process while producing better solutions. These results suggest that scalable multi-agent construction requires not only parallel decoding but also a structured mechanism that ensures simultaneous decisions are locally grounded and mutually compatible.

\subsection{Projection-Window Fidelity and Sensitivity}
\label{subsec:projection_geometry_effect}

\begin{figure}[t]
\centering
\includegraphics[width=\columnwidth]{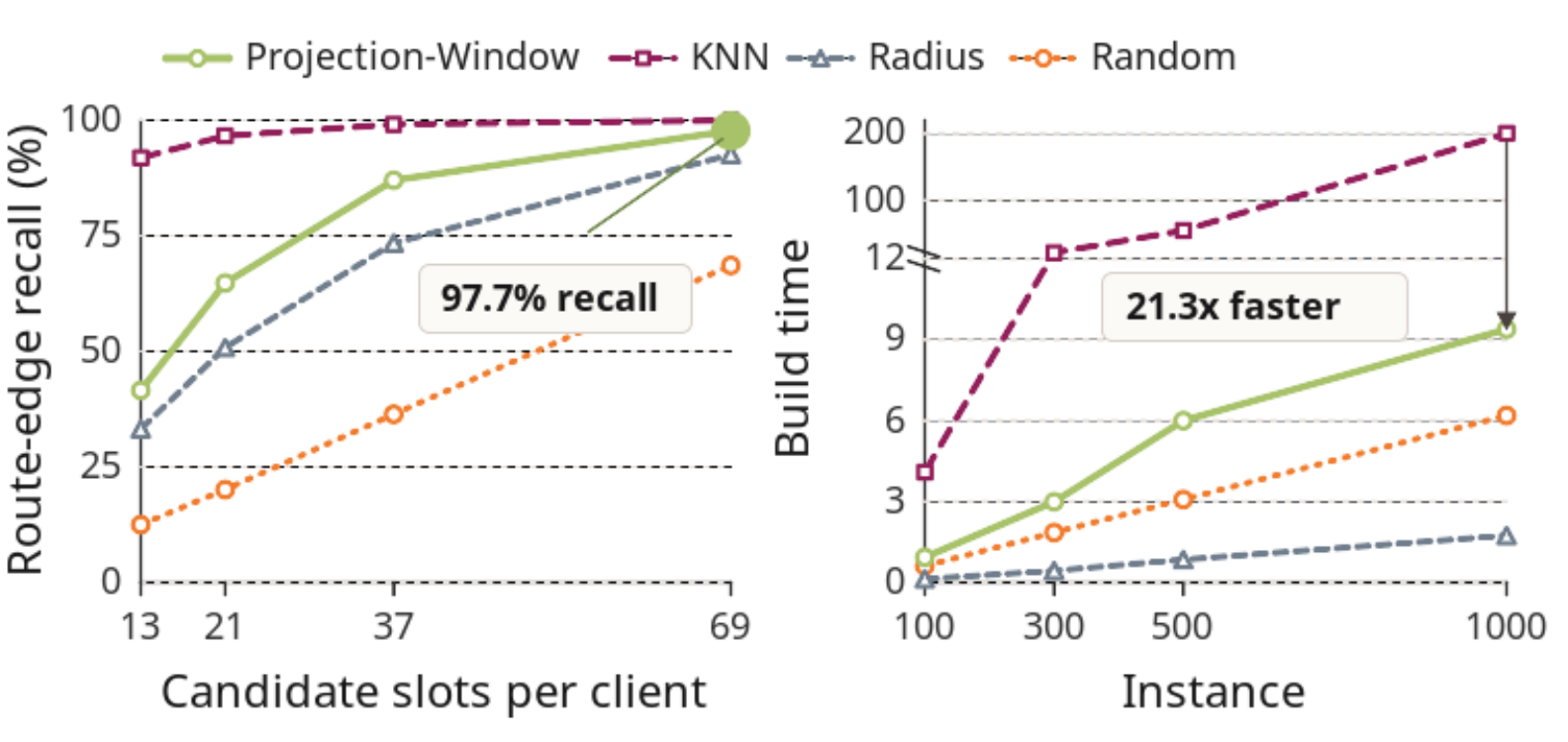}
\caption{Validation of path edges based on the geometric structure of the sliding window. Left: recall of client-to-client route transitions produced by PARCO. Right: Construction time using the same CPU. Radius denotes a radial-window baseline that sorts clients by depot distance and keeps the same local rank window.}
\label{fig:route_edge_recall}
\end{figure}

Figure~\ref{fig:route_edge_recall} evaluates whether PWin captures useful local structures or merely sparsifies the graph. We compare four candidate-set construction strategies under the same setting. Random samples candidate nodes uniformly without using geometry. Radius ranks nodes by depot distance and retains a local window in this one-dimensional order. KNN selects nearest neighbors from the complete distance matrix, which requires dense pairwise computation. Projection sorts task nodes along multiple directions and collects neighbors from the corresponding sliding rank windows, approximating local Euclidean neighborhoods without building the complete distance graph.

The left panel reports the proportion of task-to-task transitions retained by each strategy. Projection preserves most route edges, achieving recall close to distance-based KNN and clearly above Random and Radius. This suggests that multi-directional projection ordering retains relevant local successors more effectively than geometry-agnostic sampling or depot-distance ranking. The right panel reports cache-construction time. KNN has high recall, but the cost of dense pairwise distance computation grows rapidly with $N$. Projection keeps task-relevant edges in a compact candidate pool without a dense graph, providing a favorable balance between local coverage and construction cost.

\begin{table}[t]
\centering
\caption{Sensitivity analysis regarding the number of projection directions $q$. Time refers to the amortized strategy expansion time per instance (in milliseconds). Bold and underlined indicate the optimal and second-best results based on unrounded results.}
\label{tab:projection_q_sensitivity}
\small
\renewcommand{\arraystretch}{0.9}
\setlength{\tabcolsep}{0.3pt}
\begin{tabular*}{\columnwidth}{@{\extracolsep{\fill}}crrrrrrrrrrrrr@{}}
\toprule
\multirow{2}{*}{$q$} & \multirow{2}{*}{Slots} & \multicolumn{3}{c}{3000/120} & \multicolumn{3}{c}{3000/240} & \multicolumn{3}{c}{5000/200} & \multicolumn{3}{c}{5000/400} \\
\cmidrule(lr){3-5}\cmidrule(lr){6-8}\cmidrule(lr){9-11}\cmidrule(l){12-14}
 & & Obj. & Steps & Time & Obj. & Steps & Time & Obj. & Steps & Time & Obj. & Steps & Time \\
\midrule
1 & 18  & 7.08 & 172 & \textbf{23.7} & 3.95 & 141 & \textbf{36.6} & 7.19 & 213 & \textbf{71.4} & \underline{4.12} & 185 & \textbf{121.7} \\
2 & 35  & 7.06 & 165.5 & \underline{25.1} & 3.94 & 133 & \underline{36.8} & 7.19 & 213.5 & \underline{74.8} & \textbf{4.10} & 196 & 130.4 \\
4 & 69 & \underline{7.06} & \textbf{153.5} & 27.2 & \textbf{3.94} & \underline{129} & 39.1 & \underline{7.18} & \textbf{197.6} & 76.1 & 4.13 & \textbf{177.8} & \underline{125.9} \\
8 & 137 & \textbf{7.06} & \underline{157} & 32.9 & \underline{3.94} & \textbf{124.4} & 40.3 & \textbf{7.17} & \underline{205} & 90.0 & 4.15 & \underline{179.6} & 140.9 \\
\bottomrule
\end{tabular*}
\end{table}

\begin{table}[t]
\centering
\caption{Each setting reports objective and average rollout steps under greedy decoding. Lower is better.}\label{tab:hcvrp_2x2_ablation}
\small
\renewcommand{\arraystretch}{0.9}
\setlength{\tabcolsep}{0.8pt}
\begin{tabular}{@{}lccrrrrrrrr@{}}
\toprule
\multirow{2}{*}{Method} & \multirow{2}{*}{PWin} & \multirow{2}{*}{CA} & \multicolumn{2}{c}{100/7} & \multicolumn{2}{c}{1000/20} & \multicolumn{2}{c}{3000/60} & \multicolumn{2}{c}{5000/100} \\
\cmidrule(lr){4-5}\cmidrule(lr){6-7}\cmidrule(lr){8-9}\cmidrule(lr){10-11}
 &  &  & Obj. & Steps & Obj. & Steps & Obj. & Steps & Obj. & Steps \\
\midrule
PARCO & \na{} & \na{} & 4.87 & 89 & 13.97 & 693 & 16.00 & 1485 & 18.01 & 2109 \\
+CA & \na{} & \checkmark{} & 4.99 & \textbf{34} & 14.01 & \textbf{204} & 16.10 & 480 & 27.85 & 956 \\
+PWin & \checkmark{} & \na{} & \textbf{4.85} & 89 & \textbf{13.65} & 597 & 17.66 & 1292 & 21.75 & 1764 \\
GeoPAR & \checkmark{} & \checkmark{} & 4.98 & 35 & 13.79 & 215 & \textbf{13.91} & \textbf{292} & \textbf{14.16} & \textbf{353} \\
\bottomrule
\end{tabular}
\end{table}

\begin{table*}[t]
\centering
\caption{Internal ablation of CA assignment on HCVRP. Utilization is $N/(M\! \times\! \mathrm{Steps})$, and time is reported in amortized milliseconds per instance. Lower is better except for utilization. Best results are shown in bold, and second-best results are underlined.}
\label{tab:ca_internal_ablation}
\small
\renewcommand{\arraystretch}{0.9}
\setlength{\tabcolsep}{1.8pt}
\begin{tabular*}{\textwidth}{@{\extracolsep{\fill}}lrrrrrrrrrrrr@{}}
\toprule
\multirow{2}{*}{Variant} & \multicolumn{4}{c}{1000/20} & \multicolumn{4}{c}{3000/240} & \multicolumn{4}{c}{5000/400} \\
\cmidrule(lr){2-5}\cmidrule(lr){6-9}\cmidrule(l){10-13}
 & Obj. & Steps & Util. & Time & Obj. & Steps & Util. & Time & Obj. & Steps & Util. & Time \\
\midrule
PWin-only & \textbf{13.694} & 593.0 & 0.084 & 10.12 & 15.525 & 1242.0 & 0.010 & 343.38 & 24.440 & 1739.9 & 0.007 & 1204.20 \\
Logit-CF & 13.921 & 272.0 & 0.184 & 5.01 & 4.360 & 222.0 & 0.056 & 63.99 & 5.065 & 289.6 & 0.043 & 205.91 \\
GeoPool & 13.890 & 273.0 & 0.183 & 5.04 & 4.585 & 207.6 & 0.061 & 59.75 & 5.689 & 276.4 & 0.046 & 196.18 \\
GeoScore & 13.812 & 218.0 & 0.229 & \underline{4.15} & \underline{4.080} & \textbf{106.0} & \textbf{0.118} & \textbf{31.64} & \underline{4.502} & \textbf{128.0} & \textbf{0.098} & \textbf{92.84} \\
Full w/o Cache & \underline{13.778} & \textbf{201.0} & \textbf{0.249} & \textbf{3.93} & 4.142 & 149.9 & 0.084 & 43.74 & 4.639 & 209.1 & 0.060 & 149.14 \\
Full CA & 13.79 & \underline{215.0} & \underline{0.233} & 4.22 & \textbf{3.94} & \underline{129.0} & \underline{0.097} & \underline{39.06} & \textbf{4.01} & \underline{177.8} & \underline{0.070} & \underline{123.83} \\
\bottomrule
\end{tabular*}
\end{table*}

Table~\ref{tab:projection_q_sensitivity} presents a sensitivity analysis of the number of projection directions. Using one or two projection directions incurs low cache overhead but often leads to longer construction processes. Increasing the number of directions to $q=4$ consistently reduces the number of rollout steps while maintaining solution quality. Although the larger cache associated with $q=8$ occasionally provides marginal improvements in the objective value, it increases the number of raw candidate slots from 69 to 137 and introduces additional runtime overhead. Across the selected instances, the objective gap between $q=4$ and $q=8$ remains below 0.5\%, while $q=4$ reduces runtime by up to 17.2\%. We therefore adopt $q=4$ as a practical trade-off between solution quality and computational cost, rather than as the optimal configuration for every individual instance.

\subsection{Complementarity of Projection Windows and Conflict-Aware Assignment}
\label{subsec:conflict_free_assignment_effect}
Table~\ref{tab:hcvrp_2x2_ablation} shows that CA assignment primarily improves effective progress, but it does not by itself substantially improve solution quality. Using CA alone considerably reduces the number of rollout steps, which confirms that many parallel decisions are wasted by duplicate selections. However, in larger-scale settings, CA alone can lead to worse objective values, because faster progress may push agents toward suboptimal assignment patterns. In contrast, using only the projection-window geometry improves candidate quality in some settings, but it cannot effectively shorten rollouts. This indicates that high-quality local structures must also be coordinated across agents.

\subsection{Component Analysis of Cache-Guided Assignment}\label{subsec:ca_component_analysis}
Having established the complementarity of PWin and CA, we further decompose CA into its internal candidate-set construction and scoring components. All five variants use the same PWin and differ only in how joint-action candidates are sampled and ranked. PWin-only excludes conflict-aware training. Logit-CF uses only the Top-$k$ candidates selected according to the corresponding logits. GeoPool additionally introduces geometric candidates retrieved from the cache, while GeoScore further incorporates the geometric priorities of these candidates into the ranking process. Full w/o Cache disables the reuse of the geometric cache in the decoder. Table~\ref{tab:ca_internal_ablation} compares these five variants against Full CA.

The full results show that PWin-only achieves the best objective on the 1,000-node instances but generalizes poorly to denser settings. Logit-CF substantially reduces the number of rollout steps, confirming that conflict-free training is essential when multiple agents compete for mutually exclusive actions. In contrast, GeoPool alone does not consistently improve either solution quality or construction efficiency. Geometric candidates become effective only when GeoScore incorporates their geometric priorities into candidate ranking, resulting in the fewest rollout steps and the shortest runtime on the 3000- and 5000-node instances. On the 3000- and 5000-node instances, Full CA trades some efficiency for improved solution quality. These results indicate that reusing geometric candidates across the representation and assignment stages is particularly important for improving solution quality on dense instances.

\subsection{Robustness Across Problem Families, Scales, and Agent Densities}
\label{subsec:robustness_density}

\begin{figure}[t]
\centering
\includegraphics[width=\columnwidth]{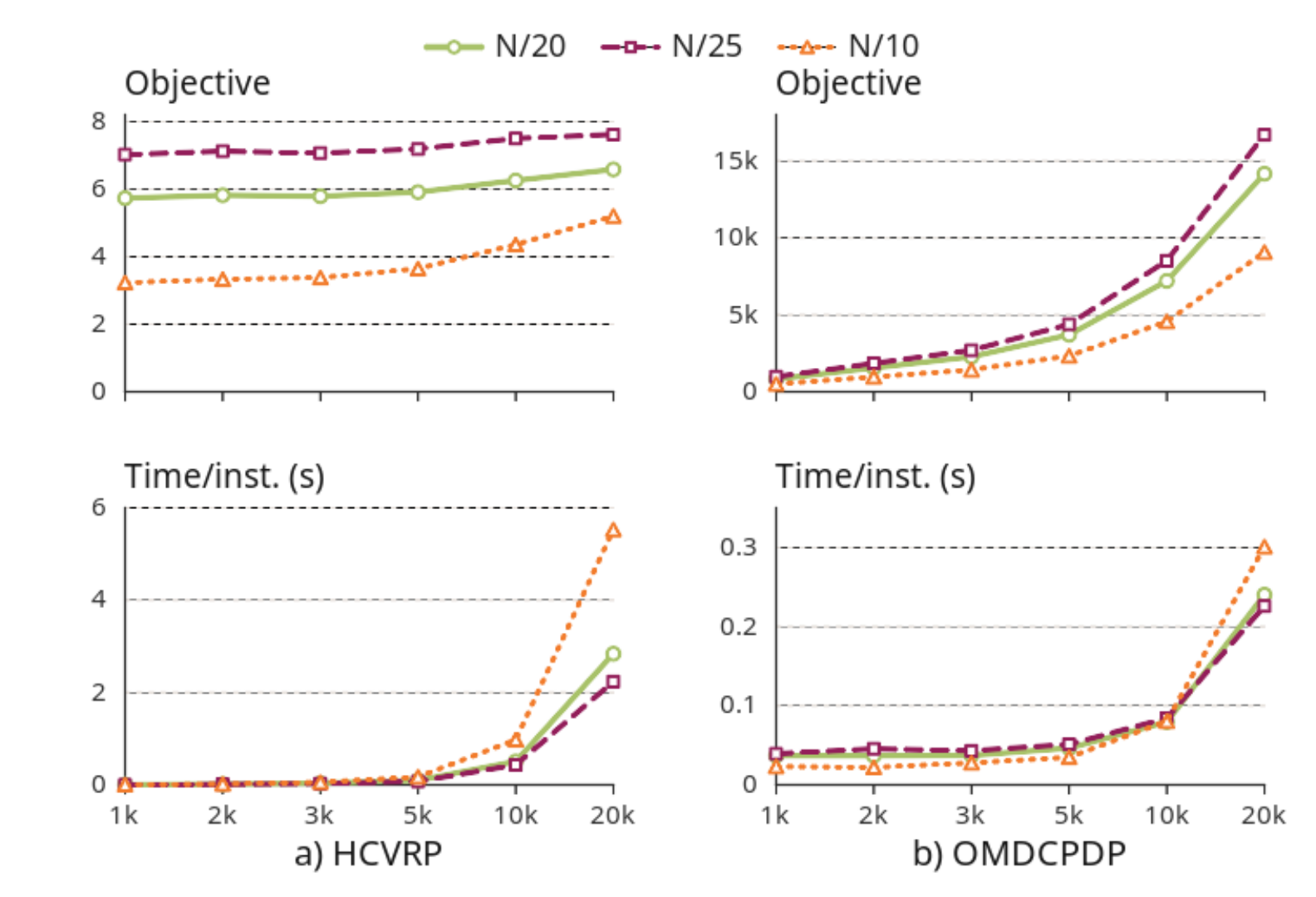}
\caption{GeoPAR objective values and per-instance inference time across problem families, problem sizes, and agent-density regimes. Lower values are better.}
\label{fig:density_robustness}
\end{figure}

Figure~\ref{fig:density_robustness} examines how GeoPAR behaves across different problem families, problem scales, and agent-density regimes. We report results for all test instances with sizes up to $N=20000$ under three agent-density settings. For HCVRP, a higher agent density reduces the min--max objective value. However, at the largest problem scale, the highest density setting, $N/10$, incurs additional runtime because parallel construction requires coordination among more agents. For OMDCPDP, the raw objective value naturally increases with $N$ because the objective accumulates delivery arrival costs over a larger number of requests. Nevertheless, the inference time per instance remains below one second even at $N=20000$.

\section{Conclusion}

We propose GeoPAR, a geometry-guided parallel autoregressive framework for large-scale multi-agent combinatorial optimization. GeoPAR builds a reusable projection-window cache, injects sparse local geometry through edge-biased attention, and reuses the cache for conflict-aware parallel assignment. Results on HCVRP and OMDCPDP show improved large-scale zero-shot generalization by preserving locally relevant actions and reducing ineffective simultaneous decisions. Ablations confirm the complementary roles of the two components: PWin improves local candidate quality, whereas CA assignment improves progress during parallel decoding.

A current limitation is that GeoPAR's improvements are uneven under extreme agent-density settings. In particular, HCVRP instances with very low agent density remain challenging because solution quality is highly sensitive to fleet scarcity, even when efficient rollout is maintained. An important future direction is to adapt candidate-action construction and assignment priorities to substantial shifts in agent density.

\bibliographystyle{ACM-Reference-Format}
\bibliography{references}

\clearpage
\def\GEOPARIncludedAppendix{}
% Standalone supplementary material for the GeoPAR arXiv preprint.
% Define \GEOPARIncludedAppendix before \input{appendix} to include only the
% appendix body in another document.
\ifdefined\GEOPARIncludedAppendix
\else
% Build from this directory with:
%   pdflatex appendix
%   pdflatex appendix
% [ADAPTED: author-visible arXiv version in the original two-column layout.]
\documentclass[sigconf,nonacm]{acmart}

\usepackage{booktabs}
\usepackage{multirow}
\usepackage{amsmath}
\usepackage{algorithm}
\usepackage{algorithmic}

\graphicspath{{figures/}}

\settopmatter{printacmref=false,printfolios=true}

\title{GeoPAR: Supplementary Material}

\author{Wenjian Wu}
\email{wjwuwwj@stu.suda.edu.cn}
\affiliation{%
  \department{School of Future Science and Engineering,}
  \institution{Soochow University}
  \city{Suzhou}
  \country{China}
}

\author{Zesheng Jia}
\email{zsjia@stu.suda.edu.cn}
\affiliation{%
  \department{School of Future Science and Engineering,}
  \institution{Soochow University}
  \city{Suzhou}
  \country{China}
}

\author{Jiaying Tang}
\email{jytang0922@stu.suda.edu.cn}
\affiliation{%
  \department{School of Future Science and Engineering,}
  \institution{Soochow University}
  \city{Suzhou}
  \country{China}
}

\author{Benyuan Yang}
\authornote{Corresponding Author.}
\email{byyang@suda.edu.cn}
\affiliation{%
  \department{School of Future Science and Engineering,}
  \institution{Soochow University}
  \city{Suzhou}
  \country{China}
}

\author{Jin Wang}
\authornotemark[1]
\email{wjin1985@suda.edu.cn}
\affiliation{%
  \department{School of Future Science and Engineering,}
  \institution{Soochow University}
  \city{Suzhou}
  \country{China}
}

\begin{document}

\maketitle
\fi

\appendix

\section*{Appendix}

\section{Problem Definitions}
\label{app:problem_definitions}

\subsection{Heterogeneous Capacitated Vehicle Routing}
\label{app:hcvrp_definition}

\paragraph{Instance.}
The min-max HCVRP consists of a depot, $N$ customer nodes, and $M$ heterogeneous vehicles. Customer $i$ has a coordinate $x_i\in[0,1]^2$ and a demand $d_i>0$. Vehicle $m$ has an initial depot location, capacity $Q_m$, and speed $v_m$. A trip starts from the depot and may return to the depot to reload. The total demand served by a vehicle between two depot visits cannot exceed its capacity. Each customer must be visited exactly once by one vehicle.

\paragraph{State and feasible actions.}
At construction step $t$, each active vehicle has a current node, a remaining capacity, and an accumulated route length. Feasible customer actions are unvisited customers whose demand fits the vehicle's remaining capacity. Depot actions are feasible when the vehicle needs to reload or when the environment exposes a safe depot move. Since customers are exclusive tasks, a joint parallel action is feasible only if no two vehicles serve the same customer in the same step. Invalid capacity, duplicate-customer, and already-served actions are masked by the environment or repaired by the conflict-aware assignment rule described in Section~\ref{app:cf_assignment}.

\paragraph{Objective.}
Let $\mathcal{R}_m=(r_{m,1},\ldots,r_{m,T_m})$ denote the realized route of vehicle $m$, including depot returns. The route duration is
\begin{equation}
    C_m =
    \frac{1}{v_m}
    \sum_{t=1}^{T_m-1}
    \lVert x_{r_{m,t+1}} - x_{r_{m,t}} \rVert_2 .
\end{equation}
The objective is to minimize the makespan,
\begin{equation}
    C_{\mathrm{HCVRP}} = \max_{m\in\{1,\ldots,M\}} C_m .
\end{equation}
The policy reward is the negative objective. We report the objective value, the average number of rollout steps, and CUDA-synchronized rollout time. A rollout step is one parallel construction step in which all currently active vehicles may propose actions.

\subsection{Open Multi-Depot Capacitated Pickup and Delivery}
\label{app:omdcpdp_definition}

\paragraph{Instance.}
OMDCPDP contains $M$ vehicle agents and $N$ task nodes, where the task nodes form $N/2$ pickup-delivery pairs. Vehicle $m$ starts from its own depot coordinate and has finite carrying capacity. For pair $k$, pickup node $p_k$ must be served before delivery node $d_k$. A delivery can be selected only after its paired pickup has been picked and is carried by a feasible vehicle state. The routes are open: vehicles are not required to return to any depot after all tasks are completed.

\paragraph{State and feasible actions.}
The dynamic state records each vehicle's current node, route length, remaining capacity, carried orders, visited pickups, completed deliveries, and pickup-delivery precedence status. A pickup action is feasible when the pickup has not been visited and the vehicle has remaining capacity. A delivery action is feasible when its paired pickup is already on board and the delivery has not been completed. Waiting, depot, or current-node actions are used only as environment-safe fallbacks when no task action is available. As in HCVRP, task actions are exclusive inside a parallel step, so duplicate task proposals must be resolved before the environment transition.

\paragraph{Objective.}
The evaluation objective is the cumulative delivery-arrival cost used by the OMDCPDP environment. Let $\mathcal{D}$ be the set of delivery nodes and let $A_j$ be the accumulated route length of the vehicle that serves delivery $j$ at the moment $j$ is completed. The reported cost is
\begin{equation}
    C_{\mathrm{OMDCPDP}} =
    \sum_{j\in\mathcal{D}} A_j ,
\end{equation}
and the policy reward is its negative value. We report objective value, rollout steps, and CUDA-synchronized rollout time under the same timing convention as HCVRP.

\section{Experimental Details}
\label{app:experimental_details}
\subsection{Implementation-Specific Model Settings}
\label{app:model_details}

\paragraph{Model configuration.}
The main paper reports the embedding and projection-window parameters. Both models also use three encoder layers, eight attention heads, RMS normalization, and one agent communication layer. The geometry branch contains a single eight-head sparse cross-attention layer with a 16-dimensional edge descriptor, no dropout, and a residual scale initialized to zero.

\paragraph{Projection-window configuration.}
\label{app}
For both datasets, $z_i$ denotes the normalized two-dimensional coordinate of task $i$. The projection directions are fixed before training and shared across all training and test instances. For $q$ directions, we set
\begin{equation}
\theta_\ell=\frac{(\ell-1)\pi}{q},
\qquad
u_\ell=(\cos\theta_\ell,\sin\theta_\ell),
\qquad \ell=1,\ldots,q.
\label{eq}
\end{equation}
Replacing $u_\ell$ with $-u_\ell$ only reverses the corresponding sorted order. Since the rank window is symmetric, the two directions are equivalent, so covering $[0,\pi)$ is sufficient. The reported models use $q=4$, corresponding to ${(1,0),2^{-1/2}(1,1),(0,1),2^{-1/2}(-1,1)}$, with a window radius of $w=8$. These directions are fixed globally and are not sampled, learned, or adapted during training. For each task, the implementation stores one explicit self slot and $2w+1$ positions per direction without removing repeated indices. This produces $1+4\times(2\times8+1)=69$ raw cache entries per task. The encoder and assignment module share the cached indices, validity masks, and geometric descriptors, and the four highest-ranked feasible cache candidates are added to the assignment pool.

\paragraph{Conflict-aware assignment.}
\label{app:cf_assignment}
Each active agent is assigned a compact candidate pool containing the four highest pointer-logit candidates, four highest projection-cache candidates, and two highest savings candidates. During training, four random feasible candidates are also added.

The savings source introduces task candidates that may be absent from both the highest pointer logits and the local projection-cache neighborhood. For agent $m$ at its current location $x_t^m$ and a feasible task $j$, we compute
\begin{equation}
    H_{\mathrm{save},t}^{m,j}
    =
    \lVert x_t^m-x_0\rVert_2
    +\lVert x_j-x_0\rVert_2
    -\lVert x_t^m-x_j\rVert_2,
    \label{eq:app_savings_score}
\end{equation}
where $x_0$ is the reference depot used by the heuristic. In the reported implementation, $x_0$ is the first depot coordinate stored in the state. It corresponds to the shared depot in HCVRP and serves as a common reference depot in OMDCPDP. A large savings value indicates that connecting the agent's current location directly to task $j$ avoids a longer path through the reference depot. Let $\mathcal{F}_t^m$ denote the feasible action set after the environment mask is applied, and let $\mathcal{V}$ denote the set of task actions. The savings candidate set is defined as
\begin{equation}
    \mathrm{SavingsK}(m,s_t)
    =
    \operatorname{TopK}_{j\in\mathcal{F}_t^m\cap\mathcal{V}}
    \left(H_{\mathrm{save},t}^{m,j};K_{\mathrm{save}}\right).
    \label{eq:app_savings_candidates}
\end{equation}
This set contains at most two feasible task actions for each agent and does not include depot or waiting actions. The savings score is used only for candidate retrieval. After a savings candidate enters the pool, it is ranked by the common assignment score without any cache-derived geometric bonus. Savings candidates are used during both training and inference, while random candidates are included only during training.

For a cached task pair $(i,j)$, let $d_{ij}=\lVert x_i-x_j\rVert_2$, and let $d_{i0}$ and $d_{j0}$ denote the distances from the two tasks to the reference depot used by the cache. We further define $\bar d$ as the mean task-to-task distance over all raw cached edges in the same instance. If task $j$ appears at window offset $\delta_{ij}\in{-w,\ldots,w}$, its normalized offset is $\rho_{ij}=|\delta_{ij}|/w$. The raw cache-geometry score is
\begin{equation}
    G_{ij}
    =
    \frac{d_{i0}+d_{j0}-d_{ij}}{\bar d}
    -\frac{d_{ij}}{\bar d}
    -0.1\rho_{ij}.
    \label{eq:app_raw_geometry_score}
\end{equation}
When an agent is currently at a depot, the corresponding cache score is $G_{0j}=-d_{0j}/\bar d_0$, where $\bar d_0$ is the mean client-to-depot distance in the instance. Let $\ell_{\theta,t}^{m,j}=\log\pi_\theta(j\mid s_t,H,m)$. Feasible cache candidates are retrieved using
\begin{equation}
    R_{\mathrm{cache},t}^{m,j}
    =0.7\ell_{\theta,t}^{m,j}+0.3G_{ij}.
    \label{eq:app_cache_retrieval_score}
\end{equation}
The four highest-scoring candidates form the cache portion of the pool. In the main experiments, the geometric bonus in Eq.~(10) of the main paper is implemented as
\begin{equation}
    B_{\mathrm{geo}}^{m,j}
    =
    \begin{cases}
        R_{\mathrm{cache},t}^{m,j}, & \text{for a cache-sourced pool entry},\\
        0, & \text{for other pool sources}.
    \end{cases}
    \label{eq:app_geometry_bonus}
\end{equation}
When the same action is proposed by multiple sources, only its first occurrence in the predefined pool-source order is retained.

For the soft cost, let $\Delta t_t^{m,j}=d(x_t^m,x_j)/v_m$ be the immediate travel time, with Euclidean distance used directly when no speed feature is present. Let $L_t^m$ be the current accumulated route length. When demand and capacity features are present, define the remaining-capacity slack as $s_t^{m,j}=Q_{t,m}^{\mathrm{rem}}-d_j$. The implementation uses
\begin{equation}
\begin{aligned}
    \Omega_t^{m,j}
    ={}&
    \lambda_{\mathrm{time}}\Delta t_t^{m,j}
    +\lambda_{\mathrm{mk}}\left(L_t^m+\Delta t_t^{m,j}\right)\\
    &+\lambda_{\mathrm{cap}}
    \left(
        [-s_t^{m,j}]_+
        +\frac{0.1}{\max(s_t^{m,j},10^{-4})}
    \right).
    \label{eq:app_soft_cost}
\end{aligned}
\end{equation}
A term is omitted when its required state feature is unavailable. Hard-infeasible capacity violations are removed by the action mask before ranking, so the reciprocal slack term primarily discourages nearly capacity-saturating feasible assignments. The reported OMDCPDP checkpoint uses no additional delivery-, pickup-, or wait-specific priority adjustment.

The final score is $S_t^{m,j}=\ell_{\theta,t}^{m,j}+\lambda_{\mathrm{geo}}B_{\mathrm{geo}}^{m,j}-\Omega_t^{m,j}$. During training, agents are processed in randomized order and tasks selected by earlier agents are masked for later agents. When conflict-aware inference is enabled, each agent proposes its top three candidates; contested tasks are assigned to the highest-scoring agent, and unresolved agents take a feasible fallback action. Table~\ref{tab:app_cf_configs} lists the main dataset-specific settings.

\begin{table}
\centering
\caption{Main conflict-aware assignment settings used in the 100-epoch training runs. The main comparison uses greedy decoding unless stated otherwise.}
\label{tab:app_cf_configs}
\small
\renewcommand{\arraystretch}{0.9}
\begin{tabular}{@{}lcc@{}}
\toprule
Setting & HCVRP & OMDCPDP \\
\midrule
Top-$r$ proposals & 3 & 3 \\
Logit / cache / savings / random & 4 / 4 / 2 / 4 & 4 / 4 / 2 / 4 \\
$\lambda_{\mathrm{geo}}$ & 0.10 & 0.10 \\
$\lambda_{\mathrm{time}}/\lambda_{\mathrm{mk}}$ & 0.10 / 0.05 & 0.10 / 0.05 \\
$\lambda_{\mathrm{cap}}$ & 1.00 & 0.00 \\
Training temperature $\tau$ & 1.2 & 1.2 \\
Exploration mixture $\epsilon$ & 0.03 & 0.03 \\
Random agent order & Yes & Yes \\
Force active in training & Yes & Yes \\
Minimum agents / tasks & 8 / 300 & 5 / 20 \\
Minimum predicted conflict & 0.03 & 0.03 \\
\bottomrule
\end{tabular}
\end{table}

\subsection{Shared Training and Evaluation Protocol}

GeoPAR uses the reinforcement-learning objective described in the main paper. Training uses Adam with zero weight decay and a multi-step learning-rate schedule with milestones at epochs 80 and 95 and a decay factor of 0.1. Training instances are generated online, whereas validation and test instances are fixed. Unless a decoding ablation states otherwise, neural models use deterministic greedy decoding.

\paragraph{SISR}
Slack Induction by String Removals (SISR)~\cite{Christiaens2020SISR} is a route-destroy-and-repair heuristic for vehicle-routing problems. We use the PARCO reference setting with $\bar{c}=10$, $L_{\max}=10$, $\alpha=10^{-3}$, $\beta=10^{-2}$, initial and final temperatures $T_0=100$ and $T_f=1$, and an iteration budget of $3\times10^5N$.

\paragraph{GA}
The Genetic Algorithm (GA)~\cite{Karakatic2015GA} evolves a population of HCVRP solutions through selection, crossover, and mutation. The population size is 200, the iteration budget is $40N$, and the mutation and crossover probabilities are $P_m=0.8$ and $P_c=1$, respectively.

\paragraph{SA}
This method~\cite{Ilhan2021SA} combines temperature-controlled search with crossover and 2-opt route refinement. We use $T_0=100$, $T_f=10^{-7}$, a Markov-chain length of $L=20N$, and a cooling factor of $\alpha=0.98$.

\paragraph{AM}
The Attention Model (AM)~\cite{Kool2019Attention} constructs an HCVRP solution sequentially by first selecting a vehicle and then its next customer, with vehicle features included in the decoder context. A separate model is trained for each small-scale $N/M$ setting using 128-dimensional embeddings, three encoder layers, eight attention heads, a rollout baseline, batch size 128, 99,968 instances per epoch, 100 epochs, learning rate $10^{-4}$ with decay 0.995. Large-scale results use the $N=100$, $M=7$ checkpoint without retraining.

\paragraph{2D-Ptr}
The 2D Array Pointer network (2D-Ptr)~\cite{Liu20242DPtr} uses separate vehicle and customer encoders and selects vehicle--customer actions with a two-dimensional pointer mechanism. Each small-scale model uses 128-dimensional embeddings, three encoder layers, eight attention heads, a rollout baseline, and a batch size of 512. The actor and critic learning rates are both $10^{-4}$ and decay by a factor of 0.995. The $N=100$, $M=7$ checkpoint is used for zero-shot large-scale evaluation with greedy decoding.

\paragraph{DPN}
Decoupling Partition and Navigation (DPN)~\cite{Zheng2024DPN} separates customer partitioning from route navigation and uses permutation symmetry across agents and rotation-based positional encoding. We use 128-dimensional embeddings, six encoder layers, eight attention heads, POMO size 60, and tanh clipping 50. It is trained for 100 epochs on $N=40$--100 and $M=3$--7 with 100,000 instances per epoch, learning rate $10^{-4}$, batch size 64 with two-step gradient accumulation. Evaluation uses greedy decoding with the identity agent permutation and no test-time augmentation.

\paragraph{OR-Tools}
Google OR-Tools~\cite{Furnon2024ORTools} provides general-purpose routing and constraint-programming solvers. The OMDCPDP solver uses a global span cost coefficient of 10,000, Path Cheapest Arc initialization, and Guided Local Search for improvement. The per-instance limits are 30, 60, and 300 seconds for $N=50$, 100, and 500, respectively, and 600 seconds for $N\geq1000$.

\paragraph{HAM}
The Heterogeneous Attention Model (HAM)~\cite{Li2021HAM} distinguishes pickup and delivery roles and enforces precedence constraints during sequential construction. Our multi-agent OMDCPDP adaptation uses 128-dimensional embeddings, three encoder layers, eight attention heads, a feed-forward dimension of 512, and batch normalization. It is trained with sampling for 100 epochs on $N=50$--100 and $M=10$--50 using batch size 128, 100,000 instances per epoch, eight augmentations, learning rate $10^{-4}$, and seed 1234; validation and test decoding are greedy.

\paragraph{MAPDP}
MAPDP~\cite{Zong2022MAPDP} learns cooperative pickup-and-delivery decisions with centralized multi-agent reinforcement learning and paired contextual embeddings. We use 128-dimensional embeddings, three encoder layers, eight attention heads, one communication layer, and the original random conflict handler. A separate checkpoint is trained for each small-scale $N/M$ setting for 100 epochs using batch size 128, 100,000 instances per epoch, eight augmentations, Adam with learning rate $10^{-4}$ and zero weight decay, and learning-rate decays of 0.1 after epochs 80 and 95. Evaluation uses greedy decoding.

\paragraph{PARCO}
PARCO~\cite{Berto2025PARCO} is the common neural baseline for both tasks. It uses three encoder layers with 128-dimensional embeddings, eight attention heads, an MLP dimension of 512, RMS normalization, one agent communication layer, and a multiple-pointer decoder. Duplicate task proposals are resolved by assigning the contested action to the agent with the highest model probability. The baseline is trained for 100 epochs with Adam, learning rate $10^{-4}$, zero weight decay, batch size 128, 100,000 instances per epoch, eight augmentations, and step decays of 0.1 after epochs 80 and 95. HCVRP training samples $N=60$--100 and $M=3$--7, whereas OMDCPDP training samples $N=50$--100 and $M=10$--50; evaluation is greedy in both tasks.

\begin{table}
\centering
\caption{Training settings of the reported checkpoints.}
\label{tab:app_training_configs}
\small
\renewcommand{\arraystretch}{0.9}
\begin{tabular}{@{}lcc@{}}
\toprule
Setting & HCVRP GeoPAR & OMDCPDP GeoPAR \\
\midrule
Training epochs & 100 & 100 \\
Learning rate & $10^{-4}$ & $10^{-4}$ \\
Train data size / epoch & 100,000 & 100,000 \\
Batch size & 128 & 128 \\
Augmentations & 10 & 8 \\
Validation monitor & n60\_m5 & n100\_m20 \\
\bottomrule
\end{tabular}
\end{table}

\subsection{Dataset Details}
\label{app:dataset_details}
\label{app:hcvrp_details}
\label{app:omdcpdp_details}

\paragraph{HCVRP}
The fixed HCVRP validation and test set contains nine splits with $N\in\{60,80,100\}$ and $M\in\{3,5,7\}$. The zero-shot and density-robustness settings are those reported in the main tables and figures, and no additional training is performed for these larger instances.

\paragraph{OMDCPDP}
The OMDCPDP checkpoint uses the shared training configuration in Table~\ref{tab:app_training_configs}. The fixed validation and test set contains six splits, n50\_m5, n50\_m7, n50\_m10, n100\_m10, n100\_m15, and n100\_m20. n100\_m20 is used for checkpoint selection. The large-scale comparisons use 128 instances per setting, batch size 1, and deterministic greedy decoding.

\section{Additional Analyses}
\label{app:additional_analyses}

\subsection{Training-Stage Quality--Efficiency Trade-off}
\label{app:training_dynamics}

\begin{figure}
\centering
\includegraphics[width=\columnwidth]{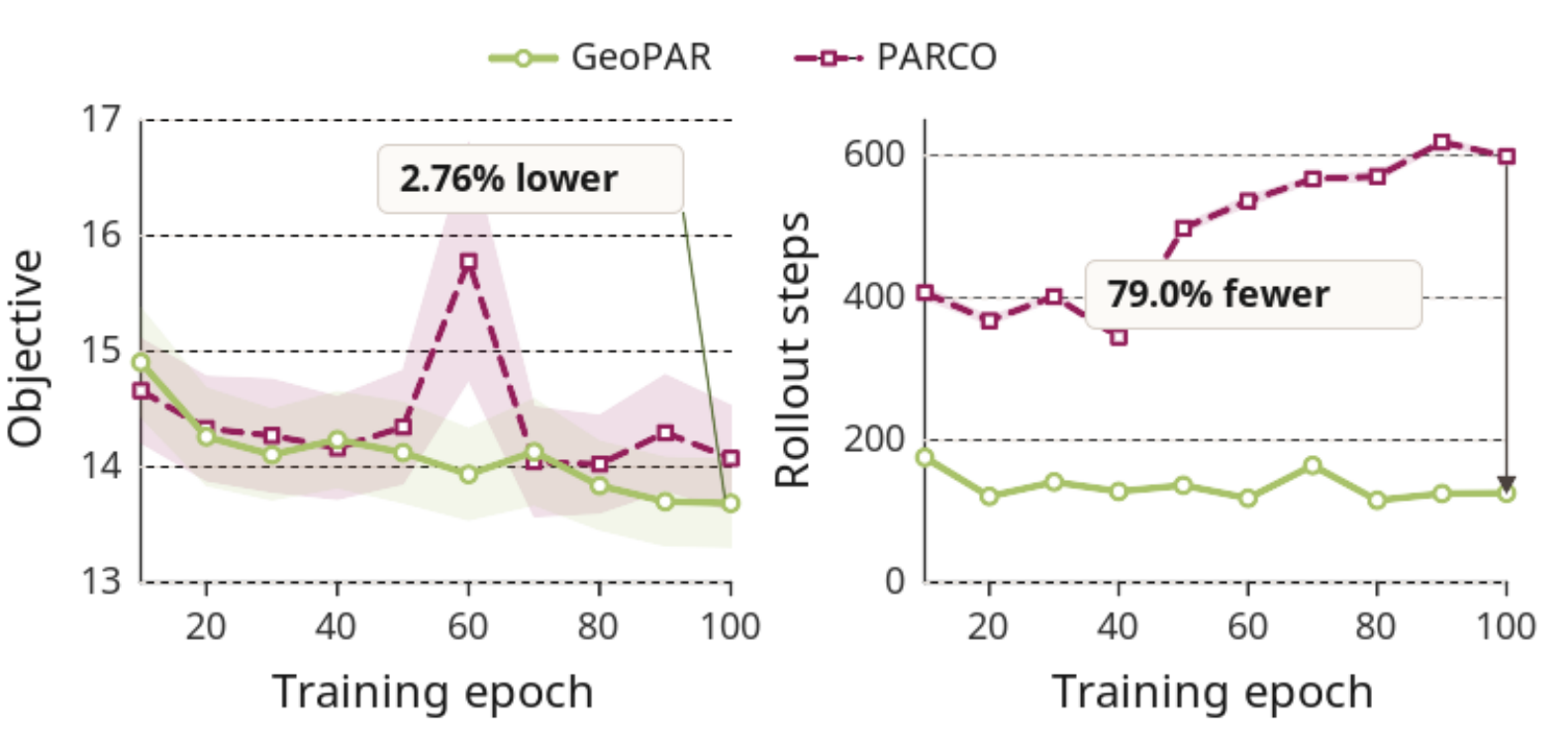}
\caption{Training dynamics of GeoPAR on 128 fixed zero-shot HCVRP instances with $N=1000$ and $M=20$. Each checkpoint is evaluated using greedy decoding with a batch size of 1. The lines and shaded bands represent the means and 95\% confidence intervals across instances.}\label{fig:app_hcvrp_training_dynamics}
\end{figure}
To distinguish the training dynamics from the final benchmark results reported in the main paper, we evaluate PARCO and GeoPAR under the same HCVRP protocol. We save checkpoints every 10 epochs and evaluate each checkpoint on a fixed zero-shot test set of 128 instances with $N=1000$ and $M=20$. Figure~\ref{fig:app_hcvrp_training_dynamics} reveals a persistent difference in the solution construction processes learned by the two models. GeoPAR requires substantially fewer rollout steps at every checkpoint, indicating that its efficiency advantage arises from the construction mechanism rather than emerging only after convergence. After 10 epochs, GeoPAR already reduces the average rollout length from 407.5 to 175.9 steps, although its objective value remains 1.67\% higher than that of PARCO. This result suggests that GeoPAR first learns a more compact decision process before fully refining solution quality. As training continues, the rollout advantage is preserved, while the objective gap is reversed. At epoch 100, GeoPAR achieves an objective value of 13.686, compared with 14.074 for PARCO, and requires only 125.6 rollout steps, compared with 599.0. The simultaneous 2.76\% improvement in objective value and 79.0\% reduction in rollout length suggest that the geometric candidate mechanism and conflict-aware assignment progressively reduce redundant and conflicting decisions, enabling the policy to improve coordination quality without relying on increasingly long construction trajectories.
\subsection{Test-Time CF Decoding Ablation}

\begin{table}
\centering
\caption{Test-time decoding ablation with standard greedy decoding or full conflict-aware assignment forced active at every step.}\label{tab:decode_cf_ablation}
\small
\renewcommand{\arraystretch}{0.9}
\setlength{\tabcolsep}{0.9pt}
\begin{tabular}{@{}lrrrrrrrrr@{}}
\toprule
\multicolumn{10}{c}{\textbf{HCVRP}} \\
\midrule
\multirow{2}{*}{Decode} & \multicolumn{3}{c}{1000/80} & \multicolumn{3}{c}{3000/240} & \multicolumn{3}{c}{5000/400} \\
\cmidrule(lr){2-4}\cmidrule(lr){5-7}\cmidrule(lr){8-10}
 & Obj. & Steps & Time & Obj. & Steps & Time & Obj. & Steps & Time \\
\midrule
Greedy & \textbf{3.81} & 69.0 & 0.663 & \textbf{3.94} & 129.0 & 1.240 & \textbf{4.13} & 177.8 & 3.981 \\
Full CA decode & 3.94 & \textbf{42.0} & \textbf{0.432} & 4.10 & \textbf{77.8} & \textbf{1.212} & 4.29 & \textbf{99.8} & \textbf{3.639} \\
\midrule
\multicolumn{10}{c}{\textbf{OMDCPDP}} \\
\midrule
\multirow{2}{*}{Decode} & \multicolumn{3}{c}{1000/25} & \multicolumn{3}{c}{3000/75} & \multicolumn{3}{c}{4000/100} \\
\cmidrule(lr){2-4}\cmidrule(lr){5-7}\cmidrule(lr){8-10}
 & Obj. & Steps & Time & Obj. & Steps & Time & Obj. & Steps & Time \\
\midrule
Greedy & 1457.2 & 44.2 & \textbf{0.061} & \textbf{3983.2} & 45.8 & \textbf{0.061} & \textbf{5205.9} & 46.2 & \textbf{0.064} \\
Full CA decode & \textbf{1452.9} & \textbf{42.0} & 0.109 & 3986.5 & \textbf{42.3} & 0.109 & 5212.0 & \textbf{42.6} & 0.111 \\
\bottomrule
\end{tabular}
\end{table}

Table~\ref{tab:decode_cf_ablation} shows that forcing full conflict-aware assignment during decoding consistently shortens rollouts in both HCVRP and OMDCPDP, but it does not necessarily improve objective values or runtime. Therefore, the benefit of conflict-aware assignment arises when it is integrated into a geometry-guided training and construction process, rather than when it is applied as an unconditional post-hoc decoding rule.

\subsection{HAM Results on Large-Scale OMDCPDP Instances}
\label{app:omdcpdp_complete_large_scale}

\begin{table}
\centering
\caption{HAM results on the large-scale zero-shot OMDCPDP settings in Table~1 of the main paper. Following the main-paper convention, the gap is computed relative to the best objective within each $N/M$ setting.}
\label{tab:app_omdcpdp_complete_large_scale}
\small
\renewcommand{\arraystretch}{0.9}
\begin{tabular*}{\columnwidth}{@{\extracolsep{\fill}}crrrr@{}}
\toprule
N/M & Obj. & Gap & Steps & Time \\
\midrule
1000/20  & 56814.07  & 3096.93\%  & 940.5  & 1.11s \\
1000/25  & 44564.27  & 2958.16\%  & 834.1  & 0.99s \\
2000/40  & 233050.98 & 6892.15\%  & 1809.6 & 2.17s \\
2000/50  & 182886.95 & 6595.99\%  & 1625.1 & 1.94s \\
3000/60  & 527148.73 & 10760.74\% & 2651.1 & 3.15s \\
3000/75  & 415019.38 & 10319.21\% & 2383.2 & 2.86s \\
4000/80  & 941895.73 & 14744.05\% & 3494.0 & 4.22s \\
4000/100 & 744548.64 & 14201.99\% & 3125.1 & 3.82s \\
\bottomrule
\end{tabular*}
\end{table}

Table~\ref{tab:app_omdcpdp_complete_large_scale} provides the previously unreported HAM results for the large-scale OMDCPDP settings included in Table 1 of the main text. Across the eight settings considered in the main comparison, the gap of HAM increases from 2958.16\% at $N=1000$ and $M=25$ to 14744.05\% at $N=4000$ and $M=80$. Over the same scale range, the number of rollout steps increases from 834.1 to 3494.0. The simultaneous deterioration in solution quality and construction efficiency indicates that the sequential construction strategy of HAM does not generalize reliably to the large-scale scenarios considered in the main comparison.

\ifdefined\GEOPARIncludedAppendix
    \def\GEOPARAppendixFinish{ }
\else
    \def\GEOPARAppendixFinish{%
        \clearpage
        \bibliographystyle{ACM-Reference-Format}
        \bibliography{references}
        \end{document}}
\fi
\GEOPARAppendixFinish

\clearpage

\end{document}